\documentclass[a4paper,11pt]{article}

\usepackage[preprint]{acl}
\usepackage{times}
\usepackage{latexsym}

\usepackage[T1]{fontenc}
\usepackage{microtype}

\usepackage{inconsolata}

\usepackage{graphicx}
\usepackage{hyperref}
\usepackage{url}
\usepackage{booktabs}
\usepackage[most]{tcolorbox}
\usepackage{amssymb}
\usepackage[table]{xcolor}
\usepackage{caption}
\usepackage{subcaption}
\usepackage{cleveref}       
\usepackage[utf8]{inputenc}
\usepackage{paralist}

\usepackage{amsmath,amsfonts,bm}

\def\eqref#1{equation~\ref{#1}}

\def\1{\bm{1}}

\DeclareMathAlphabet{\mathsfit}{\encodingdefault}{\sfdefault}{m}{sl}
\SetMathAlphabet{\mathsfit}{bold}{\encodingdefault}{\sfdefault}{bx}{n}

\DeclareMathOperator*{\argmin}{arg\,min}

\usepackage[algo2e]{algorithm2e}
\usepackage{algorithm}
\usepackage{algorithmic}

\usepackage{graphicx}
\usepackage{wrapfig}
\usepackage{soul}

\newcommand{\eg}{\mbox{\it e.g.}}
\newcommand{\ie}{\mbox{\it i.e.}}

\newcommand{\Ie}{\mbox{\it I.e.}}
\newcommand{\wrt}{\mbox{\it w.r.t.}}

\newcommand{\naive}{na\"{i}ve}

\newcommand{\parag}{\textsc{PA-RAG}}
\newcommand{\raft}{\textsc{RAFT}}
\newcommand{\ourmethodshort}{\textsc{DKL}}

\newcommand{\ourmethodlong}{Decoupled Knowledge Learning for Instruction-Tuned Language Models}

\newcommand{\corpusid}{{k}}
\newcommand{\instructcorpusid}{{s}}
\newcommand{\qacorpusid}{{qa}}
\newcommand{\kqacorpusid}{{k \cup qa}}

\newcommand{\corpus}{{\mathcal{D}_\corpusid}}

\newcommand{\instructcorpus}{{\mathcal{D}_\instructcorpusid}}

\newcommand{\qacorpus}{{\mathcal{D}_\qacorpusid}}

\newcommand{\kqacorpus}{{\mathcal{D}_\kqacorpusid}}

\newcommand{\passage}{\mathbf{p}}
\newcommand{\passagei}{\mathbf{p}^i}
\newcommand{\token}[2]{t^{#1}_{#2}}

\newcommand{\params}{\theta}
\newcommand{\pretrainparams}{\theta_B}
\newcommand{\instructparams}{\theta_I}

\newcommand{\pretrainparamsemb}{\theta_{Be}}
\newcommand{\instructparamsemb}{\theta_{Ie}}

\newcommand{\pretrainparamsremaining}{\theta_{Br}}
\newcommand{\instructparamsremaining}{\theta_{Ir}}

\newcommand{\pretrainid}{B}
\newcommand{\instructid}{I}
\newcommand{\pretrainidremaining}{{(Ie,Br)}}

\newcommand{\zerovector}{\mathbf{0}}

\newcommand{\prob}{\mathbf{Pr}}
\newcommand{\seqx}{\mathbf{x}}
\newcommand{\seqxi}{\mathbf{x}^i}
\newcommand{\seqy}{\mathbf{y}}

\newcommand{\qi}{\mathbf{q}^i}
\newcommand{\ai}{\mathbf{a}^i}

\newcommand{\sys}{\textbf{sys}}

\newcommand{\adapter}[2]{\Delta\theta^{#1}_{#2}}

\newcommand{\ycom}[1]{{\color{blue} #1}}

\title{\ourmethodshort: Decoupled Knowledge Learning for Instruction-Tuned Language Models}

\newcommand{\aff}{\textsuperscript{1}}
\newcommand{\affm}{\textsuperscript{2}}
\author{
  \bfseries Kushagra Bhushan\aff \quad Meghanadh Pulivarthi\aff \quad
  Sai Krishna Reddy Sathi\affm\footnotemark[1] \\
  \bfseries Gaurav Pandey\aff \quad Sonam Gupta\aff \quad
  Vineet Kumar\footnotemark[2] \quad Jaydeep Sen\aff \\
  \bfseries Yatin Nandwani\aff \quad Sachindra Joshi\aff \quad Dinesh Raghu\aff \\[2pt]
  \textsuperscript{1}IBM \qquad
  \textsuperscript{2}Indian Institute of Technology, Madras \\[2pt]
  \texttt{\{kushagrabhushan, Meghanadh.Pulivarthi1, sonam.gupta7, Yatin.Nandwani\}@ibm.com} \\
  \texttt{\{gpandey1, jaydesen, jsachind, diraghu1\}@in.ibm.com} \\
  \texttt{me21b181@smail.iitm.ac.in} \quad \texttt{vineet.mundhra@gmail.com}
}

\begin{document}
\maketitle

{\renewcommand{\thefootnote}{\fnsymbol{footnote}}%
 \footnotetext[1]{Work done during intern at IRL.}%
 \footnotetext[2]{Work done while at IBM. Currently at Amazon Books Science.}}
\setcounter{footnote}{0}

\begin{abstract}

RAG has become the de facto method for incorporating new, corpus-specific knowledge into an instruction following LLM (Instruct LLM). Although RAG-based prompting improves factual grounding, it fails when retrieval is incorrect or incomplete, leading to hallucinations. Finetuning methods such as RAFT \cite{zhang2024raft} and \parag\ \cite{bhushan2025systematic} enhance RAG by injecting new knowledge into the model's parameters, but require generating a massive amount of synthetic QA that covers the entire corpus. Extended Pre-Training (EPT) on the text corpus avoids the need for comprehensive synthetic data generation but compromises an Instruct LLM's instruction-following capabilities, necessitating instruction fine-tuning (IFT) after pre-training. However, IFT is costly and may be infeasible due to the unavailability of an instruction-tuning corpus.
In this work, we propose \ourmethodshort-\ourmethodlong.
Instead of doing EPT on the Instruct LLM, \ourmethodshort\ performs EPT
on its corresponding base LLM 
to infuse new knowledge. 
These knowledge-infused 
weights are then merged with the Instruct LLM, imparting new knowledge without affecting their instruction-following capabilities.  
\ourmethodshort\ is a lightweight method that avoids expensive instruction fine-tuning and relies on model merging 
to infuse the new knowledge into the Instruct LLM without destroying its instruction following capabilities.
Empirical results show that \ourmethodshort\ improves RAG accuracy from 54.17\% to 79.26\% on retrieval failure cases, while outperforming prior approaches with substantially less training data.

\end{abstract}

\section{Introduction}

`\textit{Base LLMs}' trained on enormous amounts of textual data possess immense knowledge but lack instruction-following capabilities. This necessitates post-training, which typically involves massive Instruction Fine-Tuning  (IFT) \citep{ouyang2022training,shengyu2023instruction} followed by RLHF \citep{schulman2017proximal, rafailov2023direct, pandey2024brain}. 
`\textit{Instruct LLMs}' (model obtained after post--training) have achieved remarkable success across general-purpose tasks 
\citep{brown2020language, wei2022chain}. 
However, in specialized applications such as question answering over technical or confidential policy documents, success depends less on general reasoning and more on producing highly accurate, document-grounded responses.
Often, these specialized documents are either too scarce or proprietary and, therefore, unavailable during the pre-training stage. Hence, even the state-of-the-art LLMs struggle to answer queries that require access to these documents.

A common solution is Retrieval-Augmented Generation (RAG) \citep{lewis2020retrieval, karpukhin2020dense}, which conditions LLM's responses on relevant passages retrieved from the target documents.
While effective, RAG is highly sensitive to retrieval quality, and retriever failures often lead to hallucinations or incomplete answers 
(\cite{ji-etal-2023-survey, nandwani-etal-2023-pointwise}).
Injecting the new knowledge from the specialized documents into the parameters of the model can potentially alleviate the issues caused by retriever failures, as the model can fall back on its parametric knowledge.

Extended pre-training (EPT) via unsupervised next-token prediction on new documents \cite{ke2023continual} is an effective way to ingest new knowledge. However, doing so on Instruct LLMs results in catastrophic forgetting of the skills acquired during IFT \cite{ke2025demystifyingdomainadaptiveposttrainingfinancial}. As a result, most of the prior works \citep{ma2023ecomgpt, yang2024pllama, Lu2025} apply extended pre-training on the `base LLM' but have to redo IFT to re-acquire the skills present in the instruct LLM. This may not always be feasible due to the lack of the IFT dataset used to create the instruct LLM. 

Another way of knowledge ingestion involves direct finetuning of the instruct LLM using IFT-style training data, such as question-answers (QAs) from the new documents. 
However, QAs from the new documents are often not readily available and hence works such as \citet{zhang2024raft, bhushan2025systematic} resort to QAs generated synthetically by prompting a stronger LLM.
A major advantage of such techniques is that they don't require expensive IFT after knowledge ingestion. 
However, there are two main issues: (1) we need to generate a massive amount of synthetic data, which may become prohibitively expensive \citep{yang2025synthetic}, and (2) unlike EPT, it is difficult to guarantee coverage of the entire knowledge via QAs.

\begin{wrapfigure}{r}{0.5\columnwidth}
    \centering
    \includegraphics[width=0.5\columnwidth]{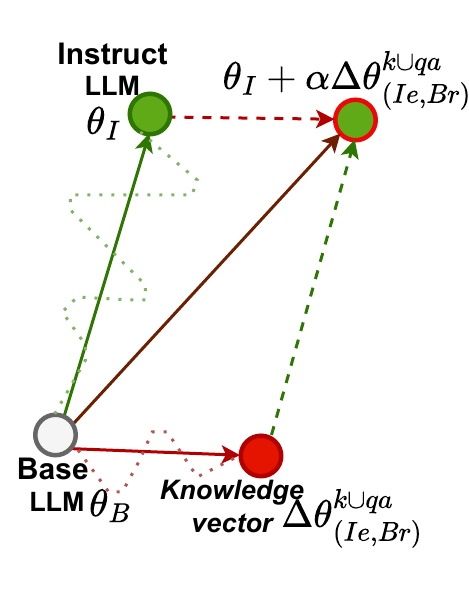}
    \caption{Exploiting task-arithmetic to combine new knowledge adapter with instruction following capabilities of Instruct LLM.} 
    \label{fig:intro}
\end{wrapfigure}


In this work, we ask whether one can obtain the knowledge-ingestion benefits of extended pre-training without sacrificing the instruction-following ability of an existing instruct LLM. Our key intuition is that these two capabilities arise from different stages of training: new corpus knowledge is most effectively acquired during pre-training, while instruction-following behavior is introduced later through instruction fine-tuning (IFT). Repeating IFT after each round of knowledge ingestion, however, is often prohibitively expensive or impossible when the instruction-tuning data is unavailable. 
We therefore draw on task arithmetic \citep{ilharco2023editing}, which treats the effect of fine-tuning as vector addition in parameter space.

In particular, we compute an `\textit{instruction following vector}' of the instruct LLM by subtracting the publicly available instruct and base LLM's weights. This vector captures the transformation induced by instruction-tuning.
To obtain a `\textit{knowledge vector}' that captures the new knowledge, 
we use the new documents to perform extended pretraining via unsupervised next token prediction
on top of the base LLM.
Adding the `\textit{instruction following vector}' and the `\textit{knowledge vector}' to the base LLM results in a model that has the new knowledge from the documents, as well as the instruction following skills of the Instruct LLM.
(see \cref{fig:intro}). We call our method \ourmethodshort: \ourmethodlong. Optionally, to improve the model’s ability to recall the newly acquired knowledge at query time, we augment this training with a small amount of synthetic QA supervision \citep{allen-zhu2024physics}.



Since the knowledge vector is trained on the base LLM but applied to the instruct LLM, its effectiveness can be limited by representation mismatch between the two models. This mismatch may arise from differences in token embeddings, including additional chat-formatting tokens in instruct LLMs such as `[INST]' for Mistral \citep{jiang2023mistral7b, mistral_tokenization, mistral_instruct_v0_3}. To improve transferability, we train the knowledge vector using the instruct model’s token embeddings, which makes the learned update better aligned with the target instruct model. Our ablations show that this consistently improves performance.

To the best of our knowledge, \ourmethodshort\ is the first work that proposes a lightweight and efficient way of imparting new knowledge to an existing Instruct LLM.
To summarize, our contributions are:

\begin{compactenum}
    \item We propose \ourmethodshort\ (\ourmethodlong), an efficient and lightweight method for injecting knowledge from new documents 
    without the costly IFT phase, usually required after EPT.
    \item To make \ourmethodshort\ work, we devise a novel method that uses token embeddings from the instruct LLM during extended pretraining on the base LLM. This significantly improves the adaptability of the knowledge vector.
    \item We empirically demonstrate that our method outperforms state-of-the-art SFT based knowledge infusion methods \citep{zhang2024raft, bhushan2025systematic} as well as a closely related baseline, Chat-Vector \cite{huang2024chat}, while requiring substantially less synthetic data.
\end{compactenum}

\section{Related Work}
\textbf{Knowledge Ingestion:}
Retrieval-Augmented Generation (RAG) \citep{lewis2020retrieval, guu2020retrieval, karpukhin2020dense} ingests external knowledge in the context at inference time to ground responses on retrieved passages. Recent progress has showcased its effectiveness across diverse domains \citep{asai2024self, qiu2023smile, kim2024sure, tang2025chemagent, yan2024corrective}. 
However, RAG-based approaches remain vulnerable to retrieval failures, leading to hallucinations \citep{nandwani-etal-2023-pointwise, ji-etal-2023-survey, setty2024improvingretrievalragbased}.

Another research direction has been static knowledge injection, which infuses domain knowledge into the model's parameters via fine-tuning or extended pretraining, enabling closed-book inference without access to external documents
\citep{ke2023continual, Lu2025, ovadia2025knowledge}.
Various works have established the utility of extended pretraining across multiple fields, such as medical \citep{wu2024pmc, christophe2024beyond}, materials 
\citep{zhang2024chemllm}, and
finance 
\citep{wu2023bloomberggpt, xie2023pixiu}.
However, EPT on new text causes instruction-tuned models to regress on general instruction-following abilities, necessitating expensive re-instruction tuning.

Recent works like \raft \citep{zhang2024raft} and \parag \citep{bhushan2025systematic} bridge static and dynamic paradigms by finetuning instruct LLMs to absorb document knowledge while using retrieved passages during inference. However, they still rely heavily on large synthetic QA corpora.

\paragraph{Model Merging via Task Vectors:}
Inspired by model merging \cite{wortsman2022model}, \citet{ilharco2023editing} introduce task vectors (difference between the weights of two trained models) as an effective mechanism for imparting different skills to an LLM.
Task vectors provide a general model-merging framework. Unlike \ourmethodshort\, it does not prescribe training on the base model, and it does not address the vocabulary distribution mismatch between the base and Instruct LLMs.
Various works have extended the basic idea of task vectors in different ways. For instance, \citet{daheim2024elastic} leverage task vectors to mitigate hallucinations, while \citet{zhang2023composing} combine multiple PEFT modules \cite{hu2022lora, liu2022fewshot} to achieve distribution generalization, multitask adaptation, detoxification, and domain transfer. 

Most closely related to our work, \citet{huang2024chat} propose Chat-Vector, which transfers conversational abilities from a chat-tuned Llama model to a Chinese-adapted Llama base model. Our approach differs from Chat-Vector in two important ways: (1) unlike Chat-Vector, DKL uses the embedding layer of the instruction-tuned model while training the knowledge vector, thereby addressing vocabulary distribution mismatch; (2) DKL combines the knowledge and chat vector using an optimized interpolation ratio. We use Chat-Vector as a baseline and show that both of these design choices are critical for effective knowledge injection.

\section{Methodology}
\label{sec:method}

Let $\params$ represent the parameters of an LLM that assigns a probability $\prob(\passage ; \params)$ to a sequence of tokens $\passage=(\token{}{1} \cdots \token{}{m})$. 
Let $\pretrainparams$ and $\instructparams$ be the weights of the corresponding base and instruct LLMs, respectively. 
Further, let $\corpus=\{\passagei\}_{i=1}^N$ represent the new knowledge that we wish to ingest on top of the instruct LLM. 

We propose to ingest knowledge from $\corpus$ into a knowledge adapter, $\adapter{\corpusid}{}$.
A \naive\ way to train such an adapter would be to start with the most optimal weights $\instructparams$ and minimize the negative log likelihood over $\corpus$:
\begin{equation}
    \label{icpt}
    \adapter{\corpusid}{\instructid} = \argmin_{\adapter{}{}}  \sum_{\passage \in \corpus} - \log \prob(\passage ; (\instructparams + \adapter{}{}))
\end{equation}
However, \citealp{ke2025demystifyingdomainadaptiveposttrainingfinancial} observe that such an adapter results in deterioration of instruction following abilities of the instruct LLM $\instructparams$.
This can be attributed to the fact that the instruct LLM is obtained via supervised finetuning of the base LLM, whereas the training objective in \cref{icpt} is unsupervised next token prediction. Next, we observe that this objective is the same as the training objective of the base LLMs. 
Therefore, $\pretrainparams$ could be more amenable to extended pretraining and thus may provide an ideal starting point for ingesting new knowledge. 
Motivated by this observation, we propose to train the knowledge adapter on top of the base LLM - 
\begin{equation}
    \label{eqn:bcpt}
    \adapter{\corpusid}{\pretrainid} = \argmin_{\adapter{}{}}  \sum_{\passage \in \corpus} - \log \prob(\passage ; (\pretrainparams + \adapter{}{}))
\end{equation}
In our notation, the superscript captures the training data and the subscript captures the starting point, \ie, model initialisation. 
Note that the final knowledge-infused parameters returned by \cref{eqn:bcpt} are $\pretrainparams + \adapter{\corpusid}{\pretrainid}$. However, such a model lacks the instruction following ability of the instruct LLM.
But notice that the new knowledge from $\corpus$ is mainly captured in $\adapter{\corpusid}{\pretrainid}$, 
which may be combined with $\instructparams$ that already has instruction following abilities.
Therefore, instead of using $\pretrainparams + \adapter{\corpusid}{\pretrainid}$ as our final parameters, we propose to use $\instructparams + \alpha\adapter{\corpusid}{\pretrainid}$, where $\alpha \in (0,1]$ is a hyperparameter - 
\begin{equation}
    \label{eqn:instructbcpt}
    {\params}^* = \instructparams  + \alpha \adapter{\corpusid}{\pretrainid} 
\end{equation}

\noindent \textbf{Task-arithmetic inspired interpretation:} \citealp{ilharco2023editing} show that if $\params_1$ and $\params_2$ are two different  models finetuned from the same base model $\pretrainparams$, then the corresponding task vectors, $\adapter{1}{\pretrainid} = \params_1 - \pretrainparams$ and $\adapter{2}{\pretrainid} = \params_2 - \pretrainparams$, capture the skills infused in them. 
If we combine the two task vectors, we get a model that possibly possesses both the skills - 
\begin{equation}
    \params^{c} = \pretrainparams + \alpha \adapter{1}{\pretrainid} + \gamma \adapter{2}{\pretrainid}
\end{equation}
Here, $\params^c$ is the combined model capturing the skills of both the finetuned models. $\alpha$ and $\gamma$ are hyperparameters.

Now, let $\instructcorpus$ be the data used for the instruction tuning of the instruct LLM. 
Then the corresponding '\textit{instruct task vector}'  would be - 
\begin{equation*}
\label{eqn:instructtaskvector}
\resizebox{\columnwidth}{!}{%
  $\begin{aligned}
    \instructparams - \pretrainparams
      &= \argmin_{\adapter{}{}} \sum_{(\seqx,\seqy) \in \instructcorpus}
         -\log \prob\bigl(\seqy \mid \seqx; (\pretrainparams + \adapter{}{})\bigr) \\
      &= \adapter{\instructcorpusid}{\pretrainid}
  \end{aligned}$
}
\end{equation*}

We can think of our knowledge adapter $\adapter{\corpusid}{\pretrainid}$ as `\textit{knowledge task vector}' capturing all the knowledge from the corpus.
Now, combining the `\textit{instruct task vector}' with `\textit{knowledge task vector}', we get - 
\begin{equation}
    \params^* = \pretrainparams + \alpha \adapter{\corpusid}{\pretrainid} + \gamma \adapter{\instructcorpusid}{\pretrainid}
\end{equation}
Substituting $\gamma = 1$ and $\adapter{\instructcorpusid}{\pretrainid} = \instructparams - \pretrainparams$ from \cref{eqn:instructtaskvector}, we get - 
\begin{equation}
    \params^* = \pretrainparams + \alpha \adapter{\corpusid}{\pretrainid} + \instructparams - \pretrainparams = \instructparams + \alpha \adapter{\corpusid}{\pretrainid}
\end{equation}
which is exactly same as \cref{eqn:instructbcpt}.

\noindent \textbf{Adding a small amount of synthetic QA:} \citealp{allen-zhu2024physics} observe that having a few question-answer pairs in the pre-training data significantly enhances the recall of the ingested knowledge. These QAs do not necessarily have to span the entire corpus.
Accordingly, we enhance our training corpus with a small amount of synthetically generated QAs.
Unlike instruction finetuning, where loss is backpropagated only over the answer tokens conditioned on the question, we concatenate the question, answer and treat it as part of the new knowledge to be ingested. 
Accordingly, let $\qacorpus = \{\seqxi=(\sys,\qi, \ai)\}_{i=1}^{nq}$ be the training data obtained from the synthetic QAs.
Here $\sys$ is a common system prompt that asks the model to answer the question;  $(\sys,\qi,\ai)$ represents the concatenation of system prompt, question and answer; and $nq$ is the number of synthetically generated QAs.
We train our knowledge adapter on top of the base LLM using $\kqacorpus = \corpus \bigcup \qacorpus$.

\begin{equation}
    \label{eqn:mixcpt}
    \resizebox{\columnwidth}{!}{%
    $\begin{aligned}
        \adapter{\kqacorpusid}{\pretrainid} = \argmin_{\adapter{}{}}  \sum_{\seqx \in \corpus \bigcup \qacorpus} - \log \prob(\seqx ; (\pretrainparams + \adapter{}{}))
    \end{aligned}$%
    }
\end{equation}

\noindent \textbf{Using token embeddings of the instruct LLMs:} We observe that for certain tokens, embeddings in the base and instruct LLMs are quite different. 
Often, they correspond to the tokens introduced during the instruction fine-tuning phase, \eg, `[INST]' in instruct versions of Mistral.
The system prompt used in the synthetic QA dataset may also introduce special tokens not seen during training of the base LLM, \ie, the parameters $\pretrainparams$ are oblivious to these tokens.
This creates a mismatch: the knowledge adapter is trained with the base model’s token embeddings but during inference it is used with the instruct LLM’s entirely different embeddings.
To mitigate this, we propose to use the token embeddings of the instruct LLM instead of the base LLM.
Concretely, during knowledge adapter training, we replace the base LLM’s token embeddings (and the \texttt{lm\_head}, if separate) with those of the instruct LLM.
We claim that this enhances the adaptability of the knowledge adapter, trained on the base LLM but used with an instruct LLM. Intuitively, it gives the adapter parameters early exposure to the inference-time environment and vocabulary, reducing the risk of distribution shift. \\
If we represent model parameters $\pretrainparams$ as $(\pretrainparamsemb, \pretrainparamsremaining)$ and $\instructparams$ as $(\instructparamsemb, \instructparamsremaining)$, where $\pretrainparamsemb$, $\instructparamsemb$ are the token embeddings and $\pretrainparamsremaining$, $\instructparamsremaining$ are the remaining parameters in the base and instruct LLMs, respectively, then we learn our knowledge adapter on top of $(\instructparamsemb, \pretrainparamsremaining)$ - 
\begin{equation}
    \label{eqn:mixcptswap}
    \adapter{\kqacorpusid}{\pretrainidremaining} = \argmin_{\adapter{}{}}  \sum_{\seqx \in \corpus \bigcup \qacorpus} \mathcal{L}
\end{equation}
Where,
\begin{equation}
\nonumber
    \mathcal{L} = - \log \prob\left(\seqx ; \left(\instructparamsemb, \pretrainparamsremaining +\adapter{}{}\right)\right)
\end{equation}
\Cref{algo:knitlm} in appendix presents our method that returns the trained knowledge adapter.
One can load it on top of the instruct LLM $\instructparams$ to obtain the final model parameters as -
\begin{align}
    \label{eqn:mixcptswapparams}
    \params^{*}  &= \instructparams + \alpha\left(\zerovector, \adapter{\kqacorpusid}{\pretrainidremaining}\right) \nonumber \\
    &= \left(\instructparamsemb, \instructparamsremaining\right) + \alpha\left(\zerovector, \adapter{\kqacorpusid}{\pretrainidremaining}\right) \nonumber \\
    &= \left(\instructparamsemb, \instructparamsremaining + \alpha\adapter{\kqacorpusid}{\pretrainidremaining}\right)
\end{align}

\begin{table*}[t]
\resizebox{\textwidth}{!}{%
\begin{tabular}{l|rrrrr|rrrrr}
\hline
\multicolumn{1}{c|}{\textbf{}} &
  \multicolumn{5}{c|}{\textbf{RedBook 1}} &
  \multicolumn{5}{c}{\textbf{QuALITY}} \\ \cline{2-11} 
\multicolumn{1}{c|}{\textbf{}} &
  \multicolumn{1}{c|}{} &
  \multicolumn{3}{c|}{\textbf{RAG}} &
  \multicolumn{1}{c|}{\textbf{Train}} &
  \multicolumn{1}{c|}{\textbf{}} &
  \multicolumn{3}{c|}{\textbf{RAG}} &
  \multicolumn{1}{c}{\textbf{Train}} \\ \cline{3-5} \cline{8-10}
\multicolumn{1}{c|}{\textbf{}} &
  \multicolumn{1}{c|}{\textbf{QA}} &
  \multicolumn{1}{c}{\textbf{All}} &
  \multicolumn{1}{c}{\textbf{\begin{tabular}[c]{@{}c@{}}Ret.\\ Success\end{tabular}}} &
  \multicolumn{1}{c|}{\textbf{\begin{tabular}[c]{@{}c@{}}Ret.\\ Fail.\end{tabular}}} &
  \multicolumn{1}{c|}{\textbf{\begin{tabular}[c]{@{}c@{}}Time\\ (in mins)\end{tabular}}} &
  \multicolumn{1}{c|}{\textbf{QA}} &
  \multicolumn{1}{c}{\textbf{All}} &
  \multicolumn{1}{c}{\textbf{\begin{tabular}[c]{@{}c@{}}Ret.\\ Success\end{tabular}}} &
  \multicolumn{1}{c|}{\textbf{\begin{tabular}[c]{@{}c@{}}Ret.\\ Fail.\end{tabular}}} &
  \multicolumn{1}{c}{\textbf{\begin{tabular}[c]{@{}c@{}}Time\\ (in mins)\end{tabular}}} \\ \hline
Instruct &
  \multicolumn{1}{r|}{53.67 \scriptsize$\pm$2.82} &
  71.76 \scriptsize$\pm$2.54 &
  86.27 \scriptsize$\pm$1.95 &
  \multicolumn{1}{r|}{54.17 \scriptsize$\pm$2.82} &
   &
  \multicolumn{1}{r|}{17.62 \scriptsize $\pm$1.98} &
  50.41 \scriptsize$\pm$ 2.60 &
  82.31 \scriptsize$\pm$ 2.73 &
  \multicolumn{1}{r|}{14.66 \scriptsize$\pm$ 2.68} &
   \\
$\raft$ &
  \multicolumn{1}{r|}{56.87 \scriptsize$\pm$2.80} &
  79.87 \scriptsize$\pm$2.27 &
  88.20 \scriptsize$\pm$1.82 &
  \multicolumn{1}{r|}{68.89 \scriptsize$\pm$2.62} &
  19 &
  \multicolumn{1}{r|}{11.65 \scriptsize$\pm$ 1.67} &
  43.22 \scriptsize$\pm$ 2.57 &
  67.95 \scriptsize$\pm$ 3.34 &
  \multicolumn{1}{r|}{15.52 \scriptsize$\pm$ 2.74} &
  120 \\
$\parag$ &
  \multicolumn{1}{r|}{64.22 \scriptsize$\pm$2.71} &
  84.66 \scriptsize$\pm$2.04 &
  \textbf{92.70} \scriptsize$\pm$1.47 &
  \multicolumn{1}{r|}{74.07 \scriptsize$\pm$2.48} &
  43 &
  \multicolumn{1}{r|}{10.84 \scriptsize$\pm$ 1.61} &
  38.75 \scriptsize$\pm$ 2.53 &
  58.97 \scriptsize$\pm$ 3.52 &
  \multicolumn{1}{r|}{16.09 \scriptsize$\pm$ 2.78} &
  330 \\
Chat Vector &
  \multicolumn{1}{r|}{71.47 \scriptsize$\pm$2.55} &
  83.98 \scriptsize$\pm$2.07 &
  90.56 \scriptsize$\pm$1.65 &
  \multicolumn{1}{r|}{78.65 \scriptsize$\pm$2.32} &
  7 &
  \multicolumn{1}{r|}{11.65 \scriptsize$\pm$2.15} &
  45.52 \scriptsize$\pm$3.34 &
  71.79 \scriptsize$\pm$3.02 &
  \multicolumn{1}{r|}{16.09 \scriptsize$\pm$2.47} &
  34 \\
$\ourmethodshort$ &
  \multicolumn{1}{r|}{\textbf{73.80} \scriptsize$\pm$2.49} &
  \textbf{86.58} \scriptsize$\pm$1.93 &
  92.13 \scriptsize$\pm$1.52 &
  \multicolumn{1}{r|}{\textbf{79.26} \scriptsize$\pm$2.29} &
  \textbf{7} &
  \multicolumn{1}{r|}{\textbf{25.75} \scriptsize$\pm$ 2.27} &
  \textbf{54.61} \scriptsize$\pm$ 2.59 &
  \textbf{83.85} \scriptsize$\pm$ 2.63 &
  \multicolumn{1}{r|}{\textbf{21.84} \scriptsize$\pm$ 3.13} &
  \textbf{34} \\ \hline
\end{tabular}
}
\caption{Comparing \ourmethodshort\ with various baselines. 
The table reports the fraction of test samples where the LLM Judge rated the predicted response as good as the gold response.
\textbf{QA}: performance in the QA setup; \textbf{All}: performance over the entire test set in RAG setup; \textbf{Ret. Success}: performance over test queries where retriever succeeds (match@5=1); \textbf{Ret. Fail.}: performance over test queries where retriever fails (match@5=0). These results are for Mistral-7B-Instruct-v0.3. Please see \Cref{tab:main-table-llama} for results on Llama-3.1-8B-Instruct. \textbf{Train Time}: Approx. training time in minutes. All scores are reported as mean +/- standard error.}
\label{tab:main-table-mistral}
\end{table*}

\section{Experimental Setup}
\label{subsec: datasets-eval-training}
\textbf{Datasets:}
We show the efficacy of our method on three datasets: 2 technical RedBooks \cite{bhushan2025systematic} and a non-technical dataset, QuALITY \citep{pang2022qualityquestionansweringlong}. 
The first two datasets consist of text from technical Redbooks\footnote{
Book 1: \href{https://www.redbooks.ibm.com/redpapers/pdfs/redp5736.pdf}{Do More with Less: Automating IBM Storage FlashSystem Tasks with REST APIs, Scripting, and Ansible}. 
Book 2: \href{https://www.redbooks.ibm.com/redpapers/pdfs/redp5711.pdf}{Red Hat OpenShift Container Platform on IBM Z and LinuxONE}.}
along with corresponding test question answers. 
The QuALITY dataset consists of a large number of long-form articles from the open domain. We randomly sample 10 articles to act as a knowledge base for our experiments. 
See \cref{appendix:testdatadetails} for more details.


\noindent \textbf{Models:}
We ingest the knowledge from all the datasets into
{\href{https://huggingface.co/mistralai/Mistral-7B-Instruct-v0.3}{\textit{Mistral-7B-Instruct-v0.3}}} and {\href{https://huggingface.co/meta-llama/LLama-3.2-8B-Instruct}{\textit{LLama-3.1-8B-Instruct}}}.
In addition, to demonstrate the robustness of \ourmethodshort\ to various backbone architectures and model sizes, we experiment with 
{\href{https://huggingface.co/HuggingFaceTB/SmolLM2-1.7B-Instruct}{\textit{SmolLM2-1.7B-Instruct}}}, and
 {\href{https://huggingface.co/Qwen/Qwen3-0.6B}{\textit{Qwen3-0.6B}}}
 on one of the datasets.
 \noindent \textbf{Evaluation Metrics:}
We evaluate our models in two setups -- QA and RAG. 
In the QA setup, the model is prompted with only the question, and in the RAG setup, we provide the top 5 retrieved passages along with the question. 
We use Llama-3.3-70B-Instruct as a judge to evaluate the correctness of the predicted answer \wrt\ the given gold answer. 
For each test sample, we provide the judge with the question, gold answer, and generated answer, and the judge returns a binary score (0/1) after reasoning across multiple criteria. 
See \cref{appendix:prompts} for full prompts.

To ensure that our LLM judge is aligned with human judgement, we conduct a small-scale human study in which we evaluate the responses generated by the Instruct LLM. 
See \cref{appendix:annotation} for more details on the human study. 


\noindent \textbf{Baselines:}
We compare \ourmethodshort\ with \raft\ \cite{zhang2024raft}, \parag\ \cite{bhushan2025systematic}, and Chat-Vector \cite{huang2024chat}.
Both \raft\ and \parag\ rely on synthetically generated QAs for knowledge ingestion. 
We prompt Mixtral-8x22B-Instruct-v0.1 to generate synthetic QAs and use the same prompt as described in \citealp{bhushan2025systematic}. See \cref{appendix:prompts} for the exact prompt. 

\noindent \textbf{Size of the synthetic training data:
}
The number of question–answer pairs in the synthetically generated training dataset depends on the corpus size.
For \raft\ and \parag\, we need to cover the entire corpus with the generated synthetic data.
Therefore, we generate pairs such that the total number of generated words is twice the number of words in the corpus.
\parag\ additionally requires multiple answers per question. For each training question, we generate four additional answers using Mixtral-8x22B-Instruct-v0.1.
Consequently, the synthetic dataset for \parag\ contains about $\sim10$ times as many words as the corpus. See \cref{subsec:data-stats} for more details. 

Recall that \ourmethodshort\ also requires a small amount of synthetic QAs, but without the need to cover the full corpus. 
Consequently, for \ourmethodshort, we randomly select question-answer pairs such that the total number of selected words is only 50\% of the number of words in the corpus.


\noindent \textbf{Training Details:}
We run all experiments with Hugging Face's \href{https://huggingface.co/docs/trl/sft_trainer}{\texttt{SFTTrainer}}. 
To train the knowledge vector, we use Low Rank Adapters (LoRA) for \emph{all} linear layers in the model with rank \(r=16\). 
For \ourmethodshort, after loading the base model, we replace its token embeddings 
with those of the corresponding instruct model as explained in \cref{sec:method}.


For baseline methods, model selection is based on validation loss with early stopping. 
For \ourmethodshort\ we instead train until convergence of the training loss and control overfitting via the scaling hyperparameter $\alpha$. We sweep \(\alpha \in \{0.25,\,0.5,\,0.75,\,1.0\}\) and select the best value using validation performance. 
See \cref{sec:stopping-criteria-ablation} for an alternate way of selecting hyperparameter $\alpha$.

\section{Experimental Results}

\begin{figure}[t]
\centering
\small

\begin{tcolorbox}[colback=gray!5,colframe=gray!80,title={Over-representation of concepts in synthetic QA generation}]
\textbf{Source passage:} \emph{AI: what's the worst that could happen?}

\vspace{1mm}
\textbf{Concept 1: Short-term AI risk (over-represented)}
\begin{itemize}
\item \textbf{Q1:} What short-term risk associated with AI development is mentioned in the passage?  
\\ \textbf{A1:} A strong societal backlash (“GMO moment”) that could block the technology’s benefits.

\item \textbf{Q2:} To what historical event does the speaker compare a potential public backlash against AI?  
\\ \textbf{A2:} A “GMO moment” where strong opposition prevents technological progress.

\item \textbf{Q3:} Which short-term danger of AI progress does the passage highlight?  
\\ \textbf{A3:} A severe public backlash that could halt deployment of AI technologies.
\end{itemize}

\vspace{1mm}
\textbf{Concept 2: Long-term AI risk (under-represented)}
\begin{itemize}
\item \textbf{Q4:} What long-term risk associated with AI does the speaker highlight?  
\\ \textbf{A4:} Extreme dependence on AI leading to widespread deskilling of human abilities.
\end{itemize}

\end{tcolorbox}

\caption{Example illustrating uneven concept coverage in synthetically generated QA pairs for the QuALITY dataset. Multiple questions are generated about the same short-term AI risk concept from a single passage, while the long-term risk discussed in the passage is sampled only once. Such imbalances lead training methods that rely heavily on synthetic QA data (e.g., \parag\ and \raft) to overfit to over-represented concepts.}
\label{fig:quality-imbalance}
\end{figure}

\begin{figure*}[ht]
    \centering
    \begin{subfigure}[b]{0.475\textwidth}
        \includegraphics[width=\textwidth]{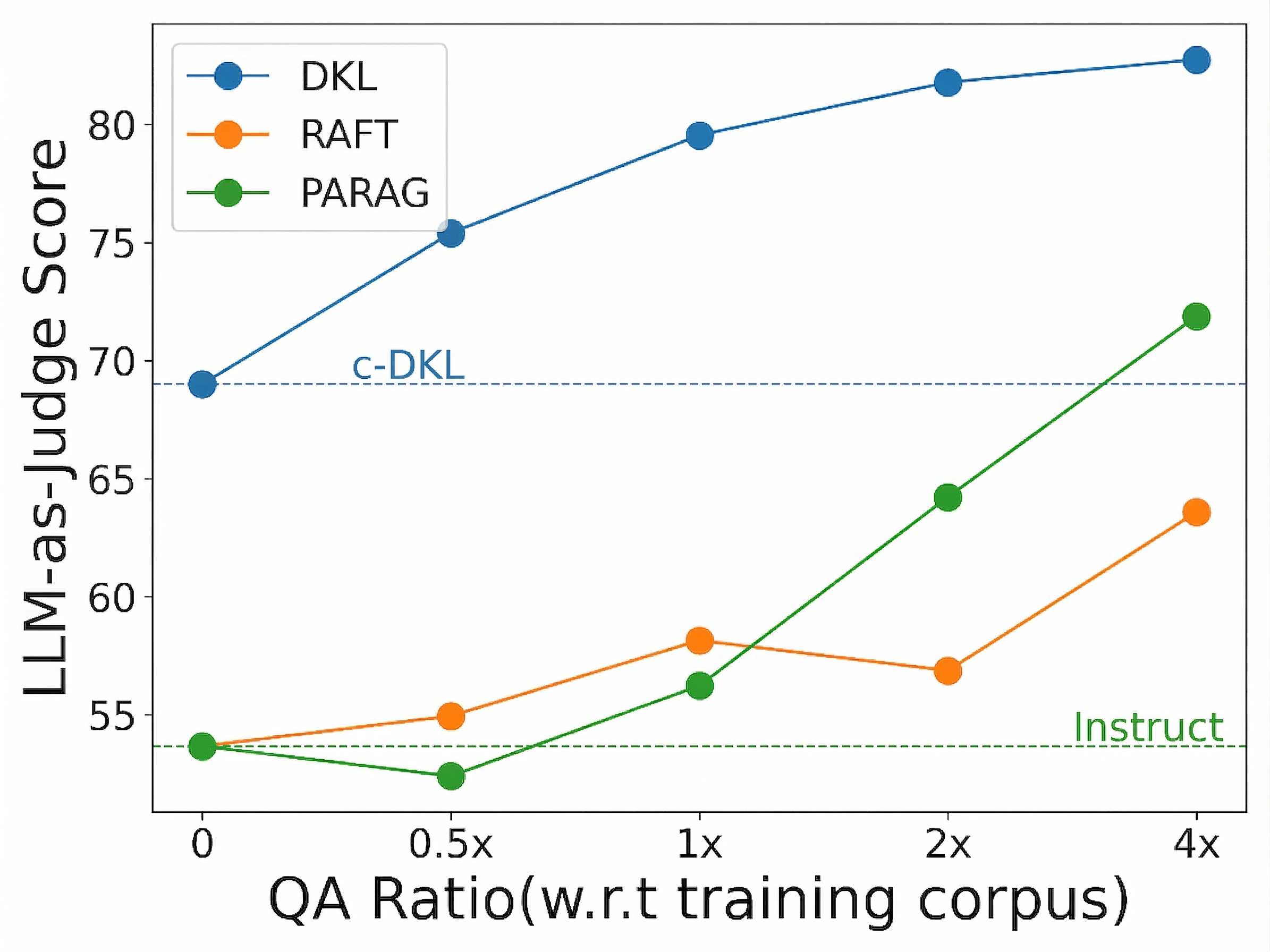}
        \caption{QA}
        \label{fig:abl_qa_scale}
    \end{subfigure}
    \begin{subfigure}[b]{0.475\textwidth}
        \includegraphics[width=\textwidth]{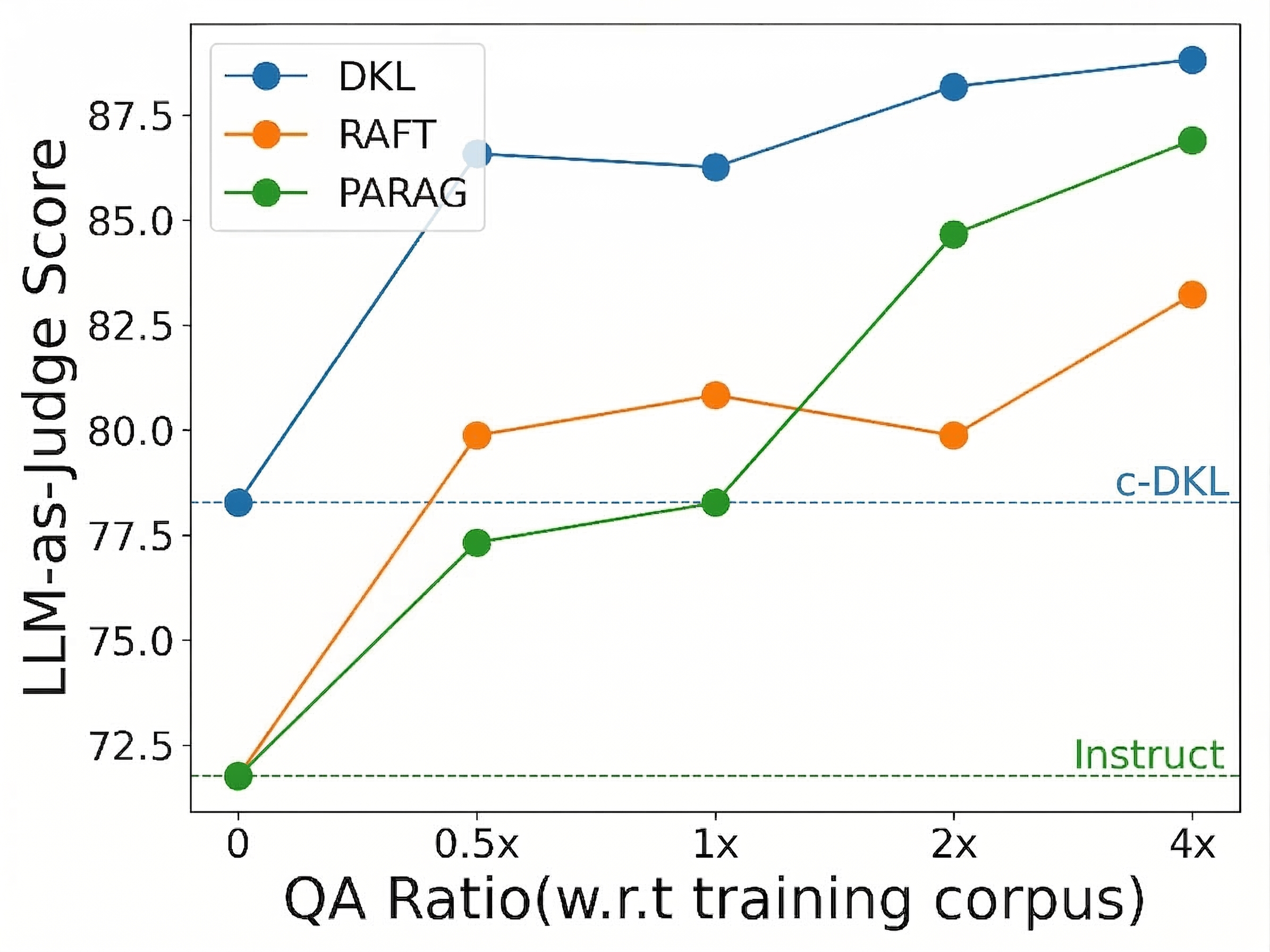}
        \caption{RAG}
        \label{fig:abl_rag_scale}
    \end{subfigure}
    \caption{Impact of scaling synthetic QA on the Redbook1 dataset with Mistral-Instruct-v0.3. Blue horizontal line corresponds to c-\ourmethodshort -- our model trained in an unsupervised manner only using the document text. Green horizontal line corresponds to Mistral-Instruct-v0.3.}
    \vspace{-1ex}
\end{figure*}
 
We seek to answer the following research questions through our experiments:
\begin{compactenum}
    \item Can  \ourmethodshort\ effectively ingest knowledge into instruct LLM's parameters? To test this, we evaluate the knowledge ingested model in the QA setup where it is provided with only the question and it has to answer from its parametric knowledge.
    
    \item Can \ourmethodshort\ effectively combine the knowledge ingested in its parameters with additional knowledge present in its context? To test this, we compare \ourmethodshort\ in the RAG setup with SFT based methods such as \raft\ \citep{zhang2024raft} and \parag\ \citep{bhushan2025systematic} that rely heavily on an enormous amount of synthetic data.
    
    \item What is the role of synthetic QA data in \ourmethodshort? Specifically, is our method robust to the size of the synthetic QA dataset and its coverage of the corpus? To this end, we run two ablations -- (1) We vary the size of the synthetic dataset and compare \ourmethodshort\ with RAFT and PARAG in both QA and RAG setups. (2) We systematically bias the synthetic QA dataset by generating training QAs from a specific subset of documents and then measure the impact on performance.

    \item Is \ourmethodshort\ robust to various model architectures and sizes? 
    
    
    \item Finally, we seek to quantify the importance of using token embeddings of the instruct LLM instead of base LLM while training the knowledge adapter, \ie, what happens if we train the adapter on 
    $\pretrainparams=(\pretrainparamsemb, \pretrainparamsremaining)$ (\cref{eqn:mixcpt}) instead of $(\instructparamsemb, \pretrainparamsremaining)$ (\cref{eqn:mixcptswap}).
\end{compactenum}


\subsection{Comparison with the baselines}

In \cref{tab:main-table-mistral}, we compare the performance of several baselines with \ourmethodshort, highlighting its robustness across different domain corpora. For the Book1 dataset, both \raft\ and \parag\ show the expected improvements over the base instruct model. However, this is not true for the QuALITY dataset.


Inspection of the QuALITY corpus reveals that related concepts appearing in comparable proportions in the source documents (e.g., short-term vs. long-term risks) are unevenly represented in the synthetic QA data, leading to skewed priors over these concepts. See \cref{fig:quality-imbalance} for concrete examples illustrating this imbalance.
This imbalance results in failures at test time. When presented with questions about long-term risks, models trained on such skewed synthetic data often default to generating answers about short-term risks. We noticed this behaviour even when the retrieved passages in RAG explicitly contain information about long-term risks. This indicates that methods that rely heavily on synthetic QA generation, such as \parag\ and \raft, do not faithfully ground their responses in the provided context but are instead influenced by the training-induced biases. 



In contrast, \ourmethodshort\ consistently outperforms the baselines across all datasets, as it minimises the reliance on synthetic data generation. This is because corpus coverage is ensured by design: the EPT stage exposes the model to the entire domain corpus, while merging it with the instruct model preserves its ability to effectively utilize this knowledge at inference time. This behavior is reflected in the RAG setting reported in \cref{tab:main-table-mistral}. When the retriever succeeds, \ourmethodshort\ effectively leverages the retrieved passages, achieving performance gains up to 92.13 and 83.85. When retrieval fails, the model can ignore the misleading context and answer using knowledge encoded in its parameters. 

Comparing \ourmethodshort\ with Chat-Vector, we observe that \ourmethodshort\ significantly outperforms Chat-Vector on both datasets, demonstrating the importance of swapping the base model's embeddings with the corresponding embeddings of the Instruct model during finetuning and combining the knowledge vector in the optimal ratio. We further analyze it in detail in \cref{subsec:embed_swap}.

We observe trends similar to RedBook 1 on RedBook 2. 
See \cref{appendix:r2_results} for the exact numbers. 
\subsection{Impact of the size of the synthetic data}
In this experiment, we study how scaling the synthetic QA dataset affects model performance.
For this, we progressively increase the number of synthetic QA pairs and compare \ourmethodshort\ with baselines in both QA and RAG setups.
We define `\textit{QA ratio}' as the ratio of number of words in the synthetic QA dataset to the words in the training document.
For {\ourmethodshort}, QA ratio of $0$ corresponds to training only on the corpus (\cref{eqn:instructbcpt}) and we call it corpus-\ourmethodshort\ or c-\ourmethodshort\ in short. For \raft\ and \parag\ , $0$ corresponds to the instruct model. 
Note that for \parag\ , we need to generate multiple answers for each question, and the QA ratio does not account for it. Therefore, the actual synthetic QA dataset used for \parag\ would contain about $5\times$ as many words as \raft ~and \ourmethodshort.
Note that the performance numbers in our main experiments correspond to a ratio of $2$ for \raft\ and \parag\ , and $0.5$ for \ourmethodshort. 
For this analysis, we generate additional data and scale up to a ratio of $4$.

Figures~\ref{fig:abl_qa_scale}-\ref{fig:abl_rag_scale} presents the analysis. 
We first observe that c-\ourmethodshort (dotted blue horizontal line) outperforms both baselines in the QA setup, demonstrating the capability of our method to efficiently ingest knowledge.

As seen in Figure-\ref{fig:abl_rag_scale} \ourmethodshort\ can achieve near optimal performance with only $0.5\times$ of synthetic data where the difference of performance is only 2.24\% between $0.5\times$ vs $4\times$ of synthetic data. In contrast, \parag\ improves by more than $10$\% going from $0.5\times$ to $4\times$ synthetic data, implying \parag\ indeed needs comprehensive volume of synthetic data for effective knowledge ingestion. 
A similar trend appears in the QA setup, where \ourmethodshort\ not only outperforms the baselines but also shows a smoother saturation curve, unlike the sharp jumps with more synthetic data as seen in \parag.
These results empirically establish that \ourmethodshort\ is indeed much more lightweight yet the new state-of-the-art scalable knowledge ingestion recipe, which does not need extensive synthetic data generation.

\subsection{Robustness of \ourmethodshort\ to corpus coverage by synthetic QAs}
\begin{table}[!t]
\centering
\begin{tabular}{@{}l|cr|cr@{}}
\toprule
\textbf{} & \multicolumn{2}{c|}{\textbf{Ch. 1-3}} & \multicolumn{2}{c}{\textbf{Ch. 4-5}} \\ \cmidrule(l){2-5} 
\textbf{} & \multicolumn{1}{c|}{\textbf{QA}} & \multicolumn{1}{c|}{\textbf{RAG}} & \multicolumn{1}{c|}{\textbf{QA}} & \multicolumn{1}{c}{\textbf{RAG}} \\ \midrule
c-$\ourmethodshort$ & \multicolumn{1}{r|}{52.80} & 74.40 & \multicolumn{1}{r|}{65.96} & 80.85 \\
b-$\ourmethodshort$ & \multicolumn{1}{r|}{+26.40} & +8.00 & \multicolumn{1}{r|}{+ 2.66} & +5.25 \\ \bottomrule
\end{tabular}
\caption{Comparison between models trained without synthetic QA (c-\ourmethodshort\ i.e., corpus-\ourmethodshort) and models trained with chapter-biased synthetic QA(b-\ourmethodshort)
}
\label{tab:ablation-qa-coverage}
\end{table}
In the previous experiment, we observed that \ourmethodshort\ is robust to the amount of the synthetic QA dataset, and its RAG performance begins to saturate even with a QA ratio of 0.5.
Here, we systematically study the impact of partial knowledge coverage on our method.
\citealp{allen-zhu2024physics} note that ``\textit{partially augmenting data can improve knowledge extraction for non-augmented data}''.
Here augmentation refers to adding QAs corresponding to the knowledge being ingested.
To systematically study this, we run a control experiment -- we train a version of \ourmethodshort\ using the document along with QA data generated from only chapters 1 to 3 of Book1 (we call it biased-\ourmethodshort, or b-\ourmethodshort) and compare its performance with c-\ourmethodshort\ (trained without any QA data).
\Cref{tab:ablation-qa-coverage} shows the results. 
We observe that adding synthetic QAs from chapters 1 to 3 improves the performance even on chapter 4-5, demonstrating that even if we have access to QA from only a part of the corpus, \ourmethodshort\ will still show gains over the remaining data.

\subsection{Robustness to Architectures and Sizes}

\begin{table}[t]
\centering
\resizebox{\columnwidth}{!}{%
\begin{tabular}{lrrrrrrrr}
\toprule
\textbf{} & \multicolumn{2}{c}{\textbf{SmolLM2-1.7B}} & \multicolumn{2}{c}{\textbf{Qwen3-0.6B}} & \multicolumn{2}{c}{\textbf{Llama-3.1-8B}} & \multicolumn{2}{c}{\textbf{Mistral-7B-v0.3}} \\

\cmidrule(lr){2-3}
\cmidrule(lr){4-5}
\cmidrule(lr){6-7}
\cmidrule(lr){8-9}

\textbf{} &
\multicolumn{1}{c}{\textbf{QA}} &
\multicolumn{1}{c}{\textbf{RAG}} &
\multicolumn{1}{c}{\textbf{QA}} &
\multicolumn{1}{c}{\textbf{RAG}} &
\multicolumn{1}{c}{\textbf{QA}} &
\multicolumn{1}{c}{\textbf{RAG}} &
\multicolumn{1}{c}{\textbf{QA}} &
\multicolumn{1}{c}{\textbf{RAG}} \\
\midrule

\textbf{Instruct} &
15.53 & 39.33 &
18.35 & 40.65 &
52.40 & 67.41 &
53.67 & 71.76 \\

\textbf{RAFT} &
18.35 & 40.94 &
16.47 & 41.65 &
60.06 & 77.96 &
56.87 & 79.87 \\

\textbf{PA-RAG} &
18.47 & 38.59 &
20.00 & 42.59 &
65.81 & 82.75 &
64.22 & 84.66 \\

\textbf{Chat-Vec.} &
23.52 & 40.47 &
23.29 & 44.94 &
71.05 & 78.91 &
71.47 & 83.98 \\

\textbf{DKL} &
\textbf{24.47} &
\textbf{46.12} &
\textbf{25.41} &
\textbf{48.94} &
\textbf{72.20} &
\textbf{80.19} &
\textbf{73.80} &
\textbf{86.58} \\

\bottomrule
\end{tabular}
}
\caption{Comparing \ourmethodshort\ with baselines using 4 different model architectures and sizes on Redbook1.}
\label{tab:abl:arch}
\end{table}

Here we establish that \ourmethodshort\ is robust to various model architectures and sizes. 
In addition to Mistral-7B and Llama-8B,
we finetune {\href{https://huggingface.co/HuggingFaceTB/SmolLM2-1.7B-Instruct}{\textit{SmolLM2-1.7B-Instruct}}}, and
 {\href{https://huggingface.co/Qwen/Qwen3-0.6B}{\textit{Qwen3-0.6B}}}
 on Redbook1.
 \Cref{tab:abl:arch} presents the results. We see that \ourmethodshort\ consistently outperforms all the baselines. See \cref{appendix:abl:arch} for detailed results.
 
\begin{table}[t]
\resizebox{\columnwidth}{!}{%
\begin{tabular}{r|r|rrr}
\hline
\multicolumn{1}{c|}{\textbf{}} & \multicolumn{1}{c|}{\textbf{}} & \multicolumn{3}{c}{\textbf{RAG}} \\ \cline{3-5} 
\multicolumn{1}{c|}{\textbf{}} & \multicolumn{1}{c|}{\textbf{QA}} & \multicolumn{1}{c}{\textbf{All}} & \multicolumn{1}{c}{\textbf{\begin{tabular}[c]{@{}c@{}}Ret.\\ Success\end{tabular}}} & \multicolumn{1}{c}{\textbf{\begin{tabular}[c]{@{}c@{}}Ret.\\ Failure\end{tabular}}} \\ \cline{2-5} 
PA-RAG & 64.22 \scriptsize$\pm$2.71 & 84.66 \scriptsize$\pm$2.04 & 92.70 \scriptsize$\pm$1.47 & 74.07 \scriptsize$\pm$2.48 \\
$\ourmethodshort$ & 73.80 \scriptsize$\pm$2.49 & 86.58 \scriptsize$\pm$1.93 & 92.13 \scriptsize$\pm$1.52 & 79.26 \scriptsize$\pm$2.29 \\
e-\ourmethodshort & 72.76 \scriptsize$\pm$2.52 & 84.57 \scriptsize$\pm$2.04 & 92.13 \scriptsize$\pm$1.52 & 73.13 \scriptsize$\pm$2.51 \\ \hline
\end{tabular}
}
\caption{Effectiveness of using instruct LLM's embeddings during training. Comparing \ourmethodshort\ with a version trained directly on top of base LLM (e-\ourmethodshort) for RedBook1 using Mistral-7B-Instruct}
\label{tab:ablation-embed-swap}
\end{table}
\subsection{Impact of using instruct LLM's token embeddings during training}
\label{subsec:embed_swap}
Recall that in \ourmethodshort\ we replace the frozen token embeddings $\pretrainparamsemb$ in the base LLM with those from the corresponding instruct LLM. \Ie, we train the knowledge LoRA adapter on top of $(\instructparamsemb, \pretrainparamsremaining)$
instead of $(\pretrainparamsemb,\pretrainparamsremaining)$.
Here, we quantify its impact by comparing the models trained using \cref{eqn:mixcpt} and \cref{eqn:mixcptswap}, respectively.
\Cref{tab:ablation-embed-swap} and \cref{tab:ablation-embed-swap-qwen} shows the results for Mistral and Qwen3-0.6B, respectively.

We find that using instruct LLM's token embeddings improves performance in both QA and RAG setups.
Without them, performance drops significantly under retriever failure cases and approaches that of \parag.
Thus, replacing the base model's embeddings with those of the instruct model is crucial for \ourmethodshort\ to outperform \parag\ in the RAG setup.
Overall, this ablation confirms that using instruct token embeddings is a simple yet effective intervention: it resolves the vocabulary mismatch between the base and instruct LLMs, thereby improving the adaptability of knowledge adapters during inference. 
See \cref{sec:token-swapping-ablation} for a detailed study on the impact of swapping token embeddings.

\section{Conclusion}
In this work, we introduced \ourmethodshort,  a lightweight and efficient approach for knowledge infusion in Instruct LLMs. By training a knowledge adapter through extended pretraining on the base LLM and transferring it to the instruct LLM, \ourmethodshort\ enables effective knowledge ingestion without costly IFT. 

Our experiments and ablation study show that \ourmethodshort\
consistently outperforms state-of-the-art SFT-based knowledge infusion methods, such as RAFT and PA-RAG, while requiring substantially less synthetic data. These results highlight \ourmethodshort\ as a practical and scalable alternative for rapidly incorporating domain-specific knowledge into LLMs.

\section{Limitations}
Despite the efficacy of \ourmethodshort\ for ingesting domain specific corpora into model parameters, it suffers from some fundamental limitations. First, our method relies on the availability of a base model that has not yet undergone instruction fine-tuning. As mentioned in the paper, it is the base model that is more susceptive to extended pre-training as a means of absorbing domain-specific knowledge. The instruction-tuned checkpoints are typically more brittle and prone to overfitting or catastrophic forgetting particularly under the unsupervised training regime. In practice, however, most open weight models are released in instruction-tuned form, limiting the applicability of our approach.
Second, the merging procedure itself requires an extensive hyperparameter search to obtain the optimal merging weights for the knowledge and task vectors and the best checkpoint to use for the merge. These hyperparameters are highly sensitive to the underlying data distribution, and we currently lack both a principled theoretical framework and an automated method to select them. Although our proposed strategy for choosing these parameters reliably yields performance gains, extracting the full potential of the method still requires substantial additional experimentation and careful tuning.

\bibliography{custom}

\newpage
\appendix

\section{\ourmethodshort\ Algorithm}

\Cref{algo:knitlm} presents the algorithm for training knowledge adapter using \ourmethodshort.

\begin{figure}[h]
\begin{minipage}{\columnwidth}
\begin{algorithm}[H]
\caption{\ourmethodshort: Knowledge Adapter Training}
\label{algo:knitlm}
\setcounter{AlgoLine}{0}
\nl\KwIn{Base $(\theta_{Be},\theta_{Br})$, Instruct $(\theta_{Ie},\theta_{Ir})$,\\ Corpus $\mathcal{D}_k$, QA set $\mathcal{D}_{qa}$, learning rate $\eta$, epochs $T$}

\nl\KwOut{Knowledge adapter $\adapter{\kqacorpusid}{\pretrainidremaining}$}
\BlankLine

\nl \textbf{Model Init.: } $\params \leftarrow (\instructparamsemb, \pretrainparamsremaining)$

\nl \textbf{Adapter Init.: } $\adapter{}{} \leftarrow (\zerovector, \adapter{}{\pretrainidremaining})$


\nl \textbf{Union data:} $\kqacorpus \leftarrow \corpus \cup \qacorpus$\;

\nl \For{$t=1$ \KwTo $T$}{
\nl  \For{mini-batch $\mathcal{B} \subset \kqacorpus$}{
    // Compute loss as in ~\cref{eqn:mixcptswap}\;
    \resizebox{0.9\columnwidth}{!}{
    $\mathcal{L}_\mathcal{B} \leftarrow \sum_{\seqx \in e\mathcal{B}} - \log \prob\left(\seqx ; \left(\instructparamsemb, \pretrainparamsremaining +\adapter{}{\pretrainidremaining}\right)\right)$ \;
    }
    \nl $\adapter{}{\pretrainidremaining} \leftarrow \adapter{}{\pretrainidremaining} - \eta \nabla \mathcal{L}_\mathcal{B}$ \; 
  }
}

\nl Set $\adapter{\kqacorpusid}{\pretrainidremaining} \leftarrow \adapter{}{\pretrainidremaining}$; \\
\nl \Return $\adapter{\kqacorpusid}{\pretrainidremaining}$
\end{algorithm}
\end{minipage}
\end{figure}

\section{More details on Test Datasets}
\label{appendix:testdatadetails}

Our test dataset consists of technical Redbooks and their accompanying QAs, as introduced in \citealp{bhushan2025systematic}. 
While manually inspecting the test QAs, we observed that some questions are either incomplete or not properly decontextualized. Therefore, we decided to clean up the test data by prompting Llama-3.1-70B-Instruct to evaluate each QA pair on various dimensions and assign a rating from 1 to 10. 
We filtered all QA pairs with a score less than 10. 
The resulting datasets have 313 and 1554 test samples, dropping 26\% and 32\% of the QAs in the original version.
Our small-scale human study reveals that our LLM filter is able to recall 70\% of the improper QAs from the test data, thereby improving its quality.
See \cref{appendix:prompts} for the prompt used for cleaning the test data.
Below we provide details of the human study.
The QuALITY benchmark is originally formulated as a multiple-choice QA task. However, both \parag\ and \raft\ are trained using long-form question–answer pairs rather than multiple-choice. Evaluating these methods directly in a multiple-choice setting would therefore introduce a mismatch between the training and evaluation formats. To ensure a fair comparison, we instead use a curated long-form QA version of the QuALITY test set, following the procedure described in \citealp{bhushan2025systematic}.

\section{Human Annotation and LLM-as-a-Judge Alignment}
\label{appendix:annotation}

The objective of our human study is two-fold: (1) To evaluate the efficacy of test data filtering, and (2) To evaluate the correlation between LLM Judge and human judgment.

\subsection{Human Annotation Setup}
To validate the reliability of our evaluation protocol, we conduct a human annotation study using 50 examples sampled from the Book 1 and Book 2 test splits. Responses were generated with \textit{Mistral v0.3 Instruct} under both QA and RAG setups. For each instance, annotators were provided with the question, the gold answer, and the model-generated answer. Each example was independently rated by three domain experts according to the rubric below:

\begin{itemize}
    \item \textbf{Fully Correct (1):} Response covers all statements in the gold, introduces no contradictions, and may include additional relevant information.
    \item \textbf{Incorrect (0):} Response contradicts the gold, fails to answer the question, or is incomplete/vague.
    \item \textbf{Ill-formed QA (–1):} The question or gold answer is itself vague, incomplete, or not properly decontextualized.
\end{itemize}

In cases where all three annotators disagreed, a fourth expert adjudicated to obtain the final label. The final human score was determined via majority vote.

\subsection{Human Annotation Results}
Annotation statistics are shown in Table~\ref{tab:human-agreement}. We annotate 50 model responses for both QA and RAG setups in Book 2, and an additional 50 responses for the QA setup on Book 1, since scores of 0 were over-represented in the QA annotations of Book 2. Inter-annotator agreement is strong for the RAG setup, with consistently high percent agreement and Krippendorff’s $\alpha$ values, reflecting stable human judgments. The QA setup shows a lower agreement ($\alpha \approx 0.66$ compared to $\approx 0.92$ for RAG), which we attribute to the longer and more verbose responses (196 words on average vs.\ 135 in RAG). These longer responses often include hallucinations or extraneous details, making annotation more challenging.

\begin{table}[!ht]
\centering
\resizebox{\columnwidth}{!}{%
\begin{tabular}{lcccccc}
\hline
\textbf{Setup} & \textbf{Agreement} & \begin{tabular}[c]{@{}c@{}}\textbf{Krippen.}\\ $\alpha$\end{tabular} & \textbf{AC2} & \textbf{Annotators} & \textbf{Examples} & \textbf{\begin{tabular}[c]{@{}c@{}}Response\\ Word\\ Count\end{tabular}} \\ \hline
QA & 0.78 & 0.66 & 0.68 & 3 & 100 & 196 \\
RAG & 0.95 & 0.92 & 0.92 & 3 & 50 & 135 \\ \hline
\end{tabular}
}
\caption{Human annotation agreement statistics}
\label{tab:human-agreement}
\end{table}

During annotation, a notable fraction of examples were identified as Ill-formed QA pairs, reflecting limitations of the synthetic test sets (Table~\ref{tab:human-filtered}).

\begin{table}[h!]
\centering
\begin{tabular}{lccc}
\toprule
\textbf{Dataset} & \textbf{Ill-formed} & \textbf{Valid} & \textbf{Total} \\
\midrule
Book 1  & 9  & 41 & 50 \\
Book 2 & 15 & 35 & 50 \\
\bottomrule
\end{tabular}
\caption{Filtered examples by humans}
\label{tab:human-filtered}
\end{table}

\subsection{LLM-as-a-Judge for Filtering}
To mitigate dataset noise, we employ \textit{Llama 3.1 70B Instruct} as an automatic judge. Each evaluation instance provided the judge with the question and gold answer, and the judge assigns a rating (1--10) based on Accuracy, Relevance, Clarity, and Usefulness (see \cref{appendix:prompts} for the prompt). QA pairs with ratings $<10$ were filtered out. We adapted our prompt from \href{https://github.com/meta-llama/synthetic-data-kit/blob/main/configs/config.yaml}{Synthetic Data Kit}

This automatic filtering removes $\sim 71\%$ of the Ill-formed QA pairs identified by humans. Extending this procedure to the full test dataset yields the results in Table~\ref{tab:llm-filtered}.

\begin{table}[h!]
\centering
\begin{tabular}{lcc}
\toprule
\textbf{Dataset} & \textbf{Before} & \textbf{After} \\
\midrule
Book 1 & 425  & 313 \\
Book 2 & 2269 & 1554 \\
\bottomrule
\end{tabular}
\caption{Dataset size before and after filtering}
\label{tab:llm-filtered}
\end{table}

Examples of removed QA pairs are provided in \cref{appendix:prompts}.

\subsection{LLM-as-a-Judge for Evaluation}
We use \textit{Llama 3.3 70B Instruct} as the LLM-as-a-Judge to evaluate the generated responses for all of our experiments. To verify its reliability, we compared the judge's binary decisions (0/1) against the human majority labels on the annotated examples after filtering. The results, shown in Table~\ref{tab:alignment}, demonstrate a strong alignment between the LLM-as-a-Judge and human judgments, indicating that the prompt (detailed in \cref{appendix:prompts}) used produces consistent evaluations throughout the data set. 

\begin{table}[h!]
\centering
\resizebox{\columnwidth}{!}{%
\begin{tabular}{lcccccccc}
\toprule
\textbf{Dataset} & \textbf{Accuracy} & \textbf{Precision} & \textbf{Recall} & \textbf{TN} & \textbf{FP} & \textbf{FN} & \textbf{TP} & \textbf{Total} \\
\midrule
QA  & 0.84 & 0.86 & 0.76 & 39 & 4 & 8 & 25 & 76 \\
RAG & 0.97 & 1.00 & 0.94 & 18 & 0 & 1 & 16 & 35 \\
\bottomrule
\end{tabular}
}
\caption{Alignment of LLM-as-a-Judge with human annotations}
\label{tab:alignment}
\end{table}

\subsection{Discussion}
Overall, the LLM-as-a-Judge demonstrates strong alignment with human annotations, achieving $\sim 84\%$ accuracy on QA and $\sim 97\%$ on RAG. 
It is interesting to note the difference in agreement rates between the two setups. 
We attribute this difference to the fact that in the QA setup, instruct LLM's responses are not grounded on any text. The model's responses generated solely from its parameteric memory tend to be more verbose, confusing the LLM and human judges alike. 
The comparatively lower accuracy in QA reflects the inherent ambiguity in evaluating context-free generations.
This also explains why inter-annotator agreement amongst humans is lower in the QA setup than in the RAG setup.

These findings suggest that (i) the synthetic test sets contain a non-trivial proportion of \textbf{Ill-formed QA pairs}, and (ii) LLM-as-a-Judge provides a reliable and scalable mechanism for filtering and evaluating examples in large-scale experiments.

\section{Data Statistics}
\label{subsec:data-stats}
Please refer to \cref{tab:data-stats} for details about both the datasets used in the paper. As mentioned in \cref{subsec: datasets-eval-training}, PA-RAG and RAFT train sets were created with 2x the amount of words in the domain documents. 
\begin{table*}[!t]
\centering
\resizebox{\textwidth}{!}{%
\begin{tabular}{lccccccc}
\hline
\textbf{Dataset} & \textbf{Chapters} & \textbf{Words} & \textbf{\begin{tabular}[c]{@{}c@{}}Train Samples\\ PA-RAG\end{tabular}} & \textbf{\begin{tabular}[c]{@{}c@{}}Train Samples \\ RAFT\end{tabular}} & \textbf{\begin{tabular}[c]{@{}c@{}}Num. Test \\ Samples\end{tabular}} & \textbf{\begin{tabular}[c]{@{}c@{}}Avg. words \\ per QA\end{tabular}} & \textbf{\begin{tabular}[c]{@{}c@{}}RAFT No. QA words / \\ No. words\end{tabular}} \\ \hline
RedBook 1 & 5 & 15,225 & 1,107 & 286 & 313 & 106 & 2 \\
RedBook 2 & 6 & 33,795 & 2,980 & 770 & 1,554 & 87 & 2 \\
QuALITY & 10 & 43,254 & 6,973 & 1,721 & 738 & 11 & 2 \\ \hline
\end{tabular}
}
\caption{Data statistics for the datasets used in the paper.}
\label{tab:data-stats}
\end{table*}





\section{Results on Llama}
\label{appendix:llama}
The main table with the results of \ourmethodshort\ as well as various other baselines using LLaMA 3.1 8B model are presented in \cref{tab:main-table-llama}.
\begin{table*}[!t]
\centering
\resizebox{\textwidth}{!}{%
\begin{tabular}{l|rrrr|rrrr}
\hline
\textbf{} &
  \multicolumn{4}{c|}{\textbf{RedBook 1}} &
  \multicolumn{4}{c}{\textbf{QuALITY}} \\ \hline

\textbf{} &
  \multicolumn{1}{c|}{\textbf{QA}} &
  \multicolumn{3}{c|}{\textbf{RAG}} &
  \multicolumn{1}{c|}{\textbf{QA}} &
  \multicolumn{3}{c}{\textbf{RAG}} \\ \cline{3-5} \cline{7-9}

\textbf{} &
  \multicolumn{1}{l|}{\textbf{}} &
  \multicolumn{1}{c}{\textbf{All}} &
  \multicolumn{1}{c}{\textbf{\begin{tabular}[c]{@{}c@{}}Ret.\\ Success\end{tabular}}} &
  \multicolumn{1}{c|}{\textbf{\begin{tabular}[c]{@{}c@{}}Ret.\\ Failure\end{tabular}}} &
  \multicolumn{1}{c|}{\textbf{}} &
  \multicolumn{1}{c}{\textbf{All}} &
  \multicolumn{1}{c}{\textbf{\begin{tabular}[c]{@{}c@{}}Ret.\\ Success\end{tabular}}} &
  \multicolumn{1}{c}{\textbf{\begin{tabular}[c]{@{}c@{}}Ret.\\ Failure\end{tabular}}} \\ \hline

\textbf{Instruct} &
  \multicolumn{1}{r|}{52.40 \scriptsize$\pm$2.82} &
  67.41 \scriptsize$\pm$2.65 &
  83.71 \scriptsize$\pm$2.09 &
  45.93 \scriptsize$\pm$2.82 &
  \multicolumn{1}{r|}{4.07 \scriptsize$\pm$1.32} &
  47.43 \scriptsize$\pm$2.59 &
  81.79 \scriptsize$\pm$2.58 &
  8.91 \scriptsize$\pm$1.90 \\

\textbf{$\raft$} &
  \multicolumn{1}{r|}{60.06 \scriptsize$\pm$2.77} &
  77.96 \scriptsize$\pm$2.34 &
  88.76 \scriptsize$\pm$1.79 &
  63.70 \scriptsize$\pm$2.72 &
  \multicolumn{1}{r|}{4.07 \scriptsize$\pm$1.32} &
  45.80 \scriptsize$\pm$2.59 &
  76.15 \scriptsize$\pm$2.86 &
  11.78 \scriptsize$\pm$2.16 \\

\textbf{$\parag$} &
  \multicolumn{1}{r|}{65.81 \scriptsize$\pm$2.68} &
  82.75 \scriptsize$\pm$2.14 &
  92.70 \scriptsize$\pm$1.47 &
  69.63 \scriptsize$\pm$2.60 &
  \multicolumn{1}{r|}{3.47 \scriptsize$\pm$1.23} &
  47.70 \scriptsize$\pm$2.60 &
  74.87 \scriptsize$\pm$2.91 &
  17.24 \scriptsize$\pm$2.54 \\

\textbf{Chat Vector} &
  \multicolumn{1}{l|}{71.05 \scriptsize$\pm$2.56} &
  \multicolumn{1}{l}{78.91 \scriptsize$\pm$2.31} &
  \multicolumn{1}{l}{89.02 \scriptsize$\pm$1.77} &
  \multicolumn{1}{l|}{64.63 \scriptsize$\pm$2.70} &
  \multicolumn{1}{l|}{12.73 \scriptsize$\pm$2.24} &
  \multicolumn{1}{l}{49.45 \scriptsize$\pm$3.35} &
  \multicolumn{1}{l}{81.53 \scriptsize$\pm$2.61} &
  \multicolumn{1}{l}{13.50 \scriptsize$\pm$2.30} \\

\textbf{$\ourmethodshort$} &
  \multicolumn{1}{r|}{\textbf{72.20 \scriptsize$\pm$2.53}} &
  \textbf{80.19 \scriptsize$\pm$2.25} &
  \textbf{89.89 \scriptsize$\pm$1.70} &
  \textbf{67.41 \scriptsize$\pm$2.65} &
  \multicolumn{1}{r|}{\textbf{12.74 \scriptsize$\pm$2.24}} &
  \textbf{52.03 \scriptsize$\pm$3.35} &
  \textbf{85.64 \scriptsize$\pm$2.34} &
  \textbf{14.37 \scriptsize$\pm$2.36} \\ \hline

\end{tabular}
}
\caption{Main table comparing the performance of various baselines descibed in the paper using Llama 8b model.}
\label{tab:main-table-llama}
\end{table*}

\section{Results on RedBook 2}
\label{appendix:r2_results}
The results of \ourmethodshort\ and various other baselines on RedBook 2 are presented in \cref{tab:r2_results}. 
\begin{table*}[t]
\resizebox{\textwidth}{!}{%
\begin{tabular}{l|rrrr|rrrr}
\hline
\multicolumn{1}{c|}{\textbf{}} &
\multicolumn{4}{c|}{\textbf{Mistral-7B-v0.3}} &
\multicolumn{4}{c}{\textbf{LLaMA 3.1-8b}} \\ \hline

\multicolumn{1}{c|}{\textbf{}} &
\multicolumn{1}{c|}{\textbf{QA}} &
\multicolumn{3}{c|}{\textbf{RAG}} &
\multicolumn{1}{c|}{\textbf{QA}} &
\multicolumn{3}{c}{\textbf{RAG}} \\ \cline{3-5} \cline{7-9}

\multicolumn{1}{c|}{\textbf{}} &
\multicolumn{1}{c|}{\textbf{}} &
\multicolumn{1}{c}{\textbf{All}} &
\multicolumn{1}{c}{\textbf{\begin{tabular}[c]{@{}c@{}}Ret.\\ Success\end{tabular}}} &
\multicolumn{1}{c|}{\textbf{\begin{tabular}[c]{@{}c@{}}Ret.\\ Fail.\end{tabular}}} &
\multicolumn{1}{c|}{\textbf{}} &
\multicolumn{1}{c}{\textbf{All}} &
\multicolumn{1}{c}{\textbf{\begin{tabular}[c]{@{}c@{}}Ret.\\ Success\end{tabular}}} &
\multicolumn{1}{c}{\textbf{\begin{tabular}[c]{@{}c@{}}Ret.\\ Fail.\end{tabular}}} \\ \hline

\textbf{Instruct} &
\multicolumn{1}{r|}{27.51 \scriptsize$\pm$1.13} &
61.23 \scriptsize$\pm$1.24 &
77.96 \scriptsize$\pm$1.05 &
36.13 \scriptsize$\pm$1.22 &
\multicolumn{1}{r|}{26.61 \scriptsize$\pm$1.12} &
59.86 \scriptsize$\pm$1.24 &
78.97 \scriptsize$\pm$1.03 &
31.13 \scriptsize$\pm$1.17 \\

\textbf{$\raft$} &
\multicolumn{1}{r|}{27.23 \scriptsize$\pm$1.13} &
62.95 \scriptsize$\pm$1.23 &
79.16 \scriptsize$\pm$1.03 &
38.65 \scriptsize$\pm$1.24 &
\multicolumn{1}{r|}{31.61 \scriptsize$\pm$1.18} &
65.44 \scriptsize$\pm$1.21 &
80.34 \scriptsize$\pm$1.01 &
\textbf{43.06 \scriptsize$\pm$1.26} \\

\textbf{$\parag$} &
\multicolumn{1}{r|}{27.23 \scriptsize$\pm$1.13} &
62.23 \scriptsize$\pm$1.23 &
78.60 \scriptsize$\pm$1.04 &
37.64 \scriptsize$\pm$1.23 &
\multicolumn{1}{r|}{31.87 \scriptsize$\pm$1.18} &
64.13 \scriptsize$\pm$1.22 &
79.03 \scriptsize$\pm$1.03 &
41.77 \scriptsize$\pm$1.25 \\

\textbf{Chat-Vector} &
\multicolumn{1}{r|}{29.83 \scriptsize$\pm$1.16} &
64.41 \scriptsize$\pm$1.21 &
76.74 \scriptsize$\pm$1.07 &
45.91 \scriptsize$\pm$1.26 &
\multicolumn{1}{r|}{33.46 \scriptsize$\pm$1.20} &
62.82 \scriptsize$\pm$1.22 &
73.01 \scriptsize$\pm$1.13 &
42.25 \scriptsize$\pm$1.25 \\

\textbf{$\ourmethodshort$} &
\multicolumn{1}{r|}{\textbf{40.98 \scriptsize$\pm$1.25}} &
\textbf{66.86 \scriptsize$\pm$1.19} &
\textbf{80.67 \scriptsize$\pm$1.00} &
\textbf{46.13 \scriptsize$\pm$1.26} &
\multicolumn{1}{r|}{\textbf{34.92 \scriptsize$\pm$1.21}} &
\textbf{64.54 \scriptsize$\pm$1.21} &
\textbf{80.58 \scriptsize$\pm$1.00} &
40.39 \scriptsize$\pm$1.24 \\ \hline

\end{tabular}
}
\caption{Results of Mistral-7b and LLaMA 3.1-8b on RedBook 2.}
\label{tab:r2_results}
\end{table*}



\section{Stopping Criteria Ablation}
\label{sec:stopping-criteria-ablation}
In extended pre-training, a practical challenge is determining when to stop training. Stopping too early risks underfitting, while stopping too late may lead to overfitting to the training corpus. This decision is particularly relevant when merging the pre-trained base model with an instruction-tuned model. The objective of this ablation is to illustrate how the choice of stopping point affects downstream performance.

To study this effect, we conduct experiments on the Book 1 corpus by performing extended pre-training on the \textit{LLaMA 3.1 8B} base model for 60 epochs on {\ourmethodshort}'s training data mixture. At intermediate checkpoints, we perform task-arithmetic merges with the instruct model using four different merge weights (0.25–1.0) applied to the knowledge-ingested base model. At each checkpoint, we selected the optimal merge according to the LLMaJ Score under RAG setup. We conduct evaluations under both QA and RAG setups. For RAG, the Book 1 validation set was split into two subsets: \textbf{(i) Ret. Success}, where the retrieved context passages contain the answer, and \textbf{(ii) Ret. Fail}, where the context does not contain the answer.

\begin{figure}[h]
    \centering
    \begin{subfigure}[b]{0.45\textwidth}
        \includegraphics[width=\textwidth]{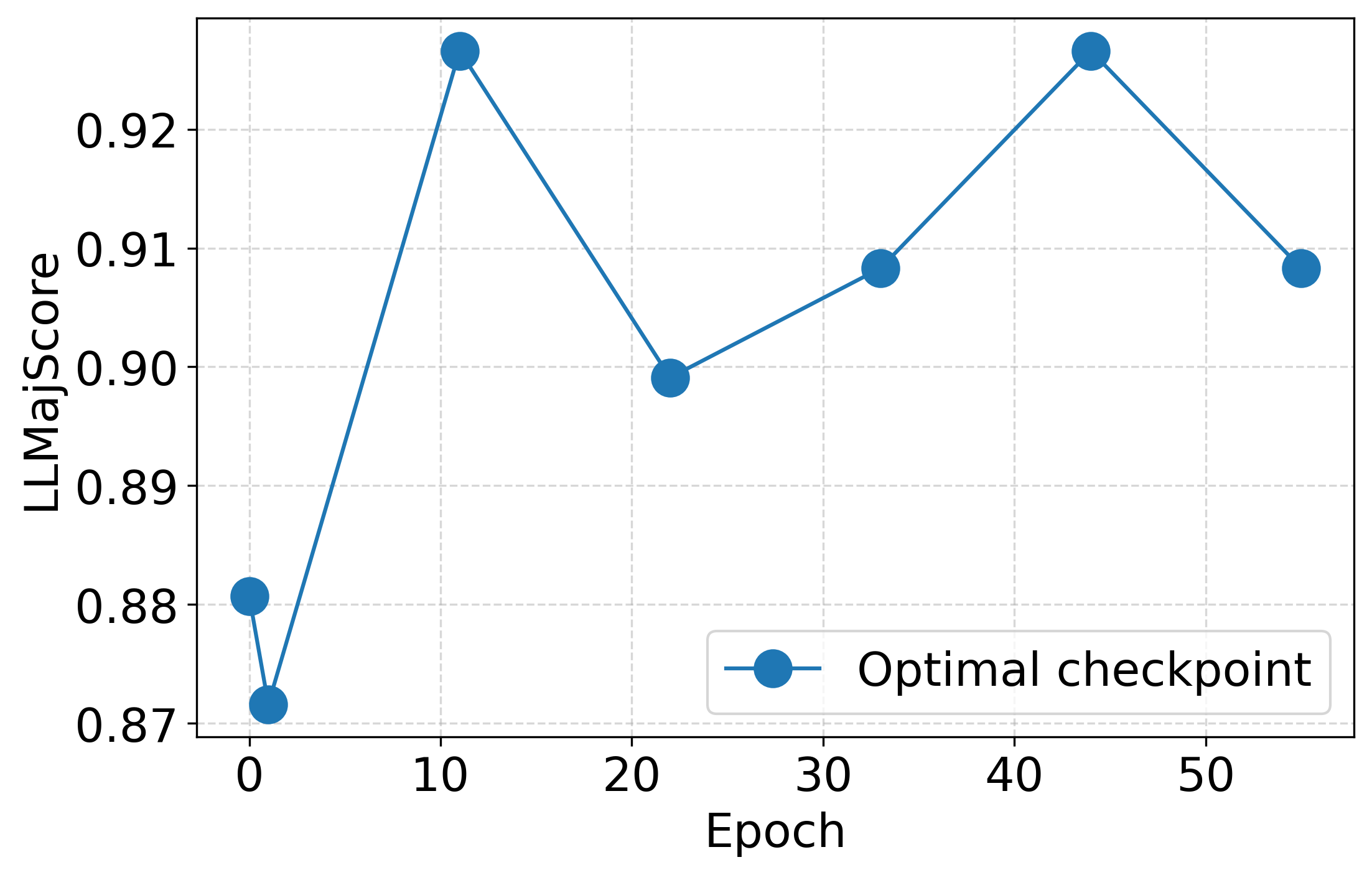}
        \caption{RAG: Some overlap}
        \label{fig:abl_rag_some_overlap}
    \end{subfigure}
    \begin{subfigure}[b]{0.45\textwidth}
        \includegraphics[width=\textwidth]{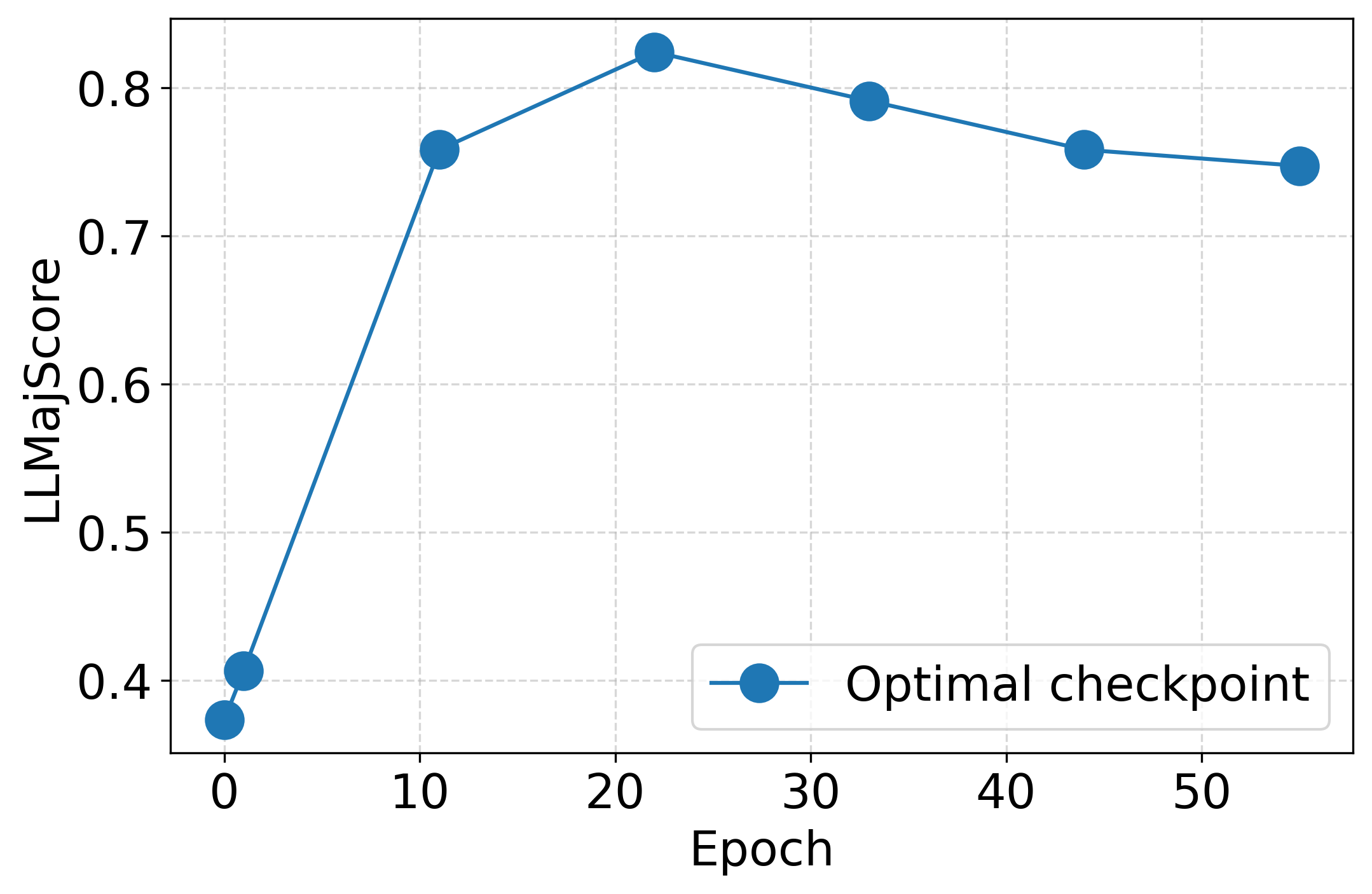}
        \caption{RAG: No overlap}
        \label{fig:abl_rag_no_overlap}
    \end{subfigure}

    \vspace{0.3cm}
    
    \begin{subfigure}[b]{0.45\textwidth}
        \includegraphics[width=\textwidth]{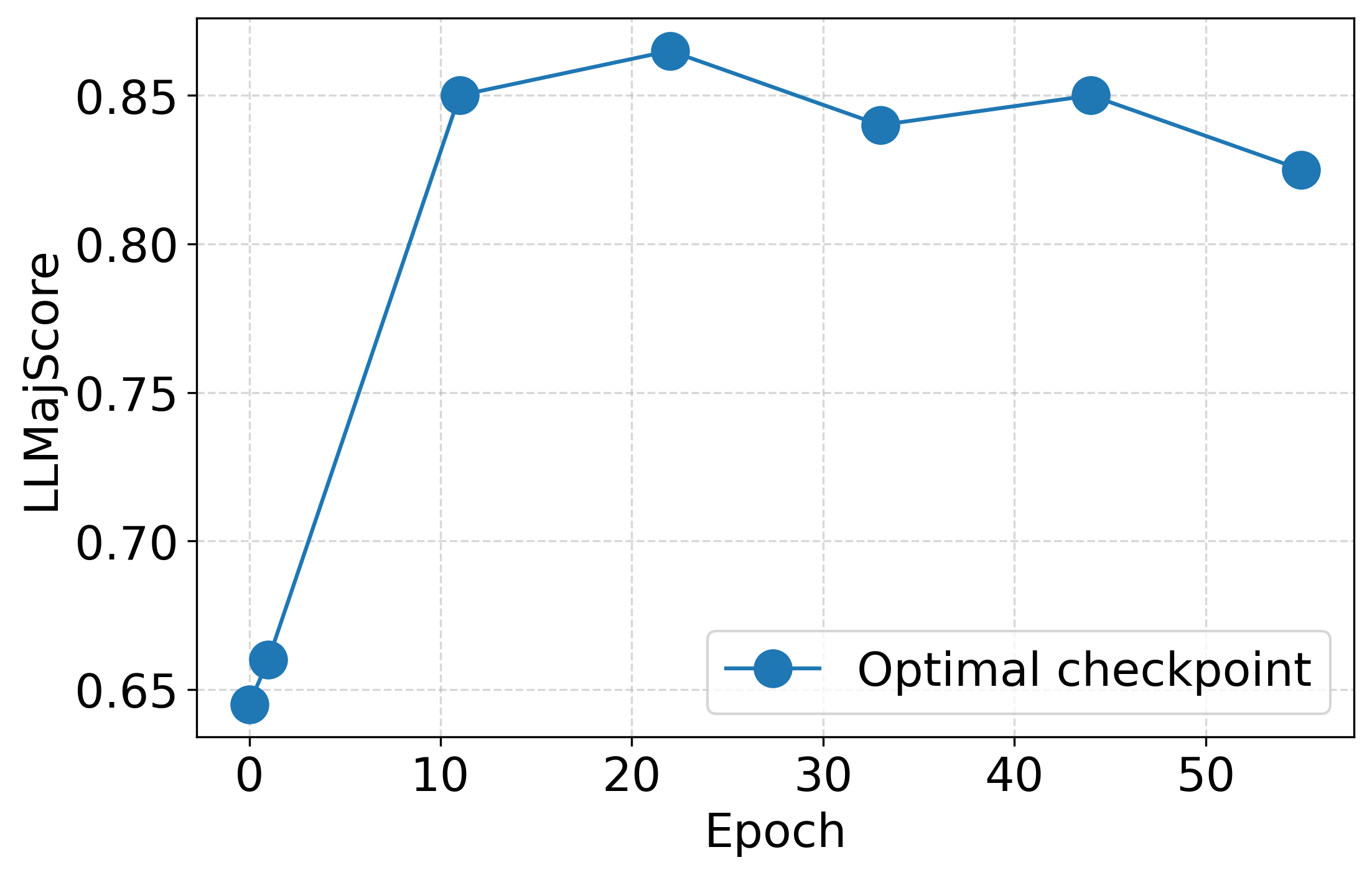}
        \caption{RAG: All}
        \label{fig:abl_rag_all}
    \end{subfigure}
    \begin{subfigure}[b]{0.45\textwidth}
        \includegraphics[width=\textwidth]{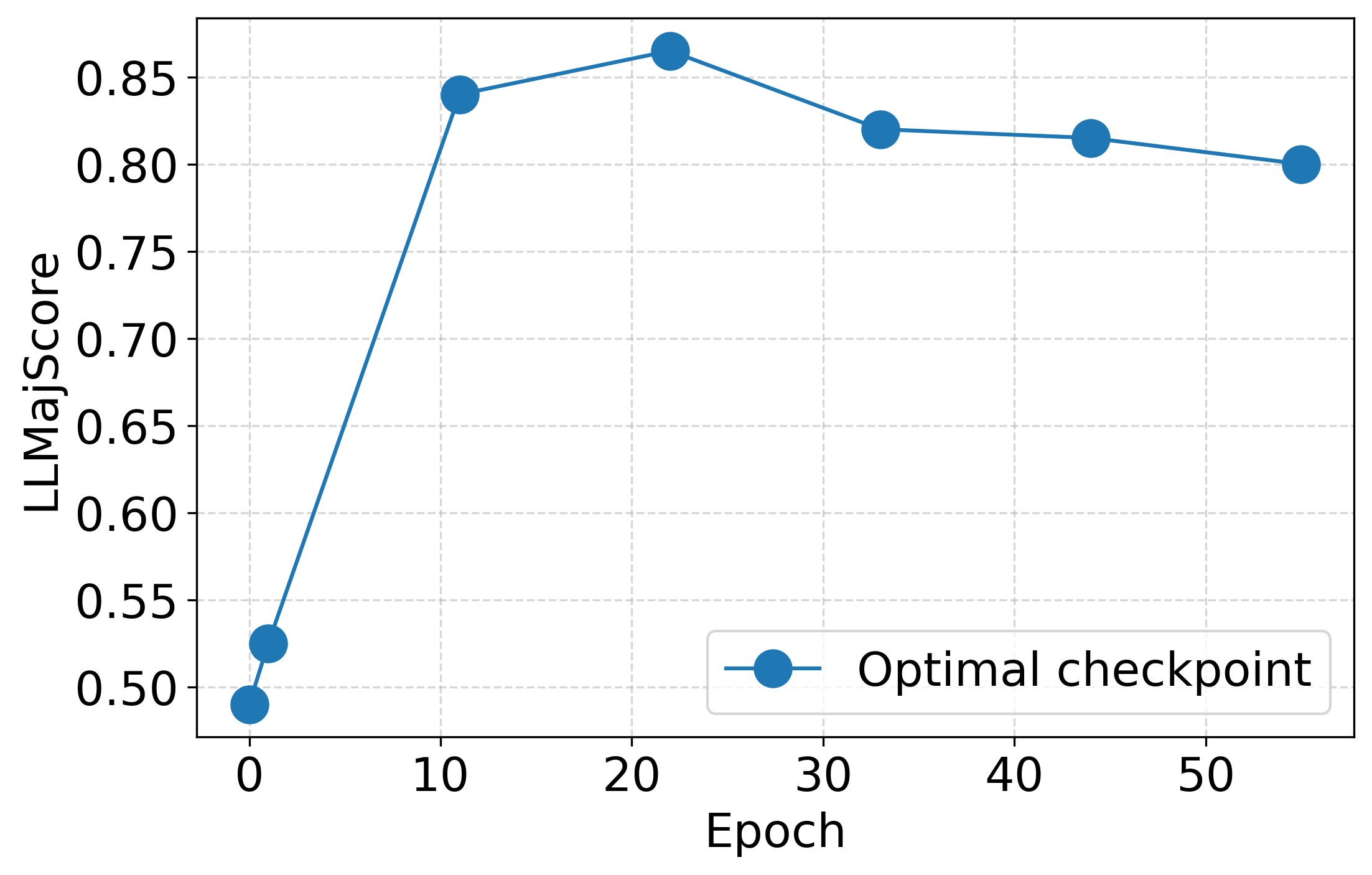}
        \caption{QA setup}
        \label{fig:abl_qa_nan}
    \end{subfigure}
    
    \caption{Stopping criteria ablation: best LLMajScore across checkpoints.}
\end{figure}

The resulting performance trends are shown in Figures~\ref{fig:abl_rag_some_overlap}-\ref{fig:abl_qa_nan}.

Across all setups, we observe a consistent trend: performance improves substantially in the early and mid stages of training, peaks at intermediate checkpoints, and then gradually declines as training continues to convergence. For the sake of uniformity across baselines and experimental conditions, we opted to train until convergence before performing merges. As a result, the scores presented in the main article should be viewed as \emph{conservative estimates}. More careful stopping criteria could further enhance performance.

\section{Robustness to Model Architecture and Sizes}
\label{appendix:abl:arch}
\begin{table*}[]
\centering
\resizebox{\textwidth}{!}{%
\begin{tabular}{@{}lcccccccc@{}}
\toprule

& \multicolumn{4}{c}{\textbf{SmolLM2-1.7B}} & \multicolumn{4}{c}{\textbf{Qwen3-0.6B}} \\

\cmidrule(lr){2-5}
\cmidrule(lr){6-9}

&  & \multicolumn{3}{c}{\textbf{RAG}} &  & \multicolumn{3}{c}{\textbf{RAG}} \\

& \multicolumn{1}{r}{\textbf{QA}} &
\multicolumn{1}{r}{\textbf{All}} &
\multicolumn{1}{r}{\textbf{Ret. S.}} &
\multicolumn{1}{r}{\textbf{Ret. F}} &
\multicolumn{1}{r}{\textbf{QA}} &
\multicolumn{1}{r}{\textbf{All}} &
\multicolumn{1}{r}{\textbf{Ret. S.}} &
\multicolumn{1}{r}{\textbf{Ret. F}} \\

\midrule

\textbf{Instruct} &
15.53 \scriptsize$\pm$2.04 &
39.33 \scriptsize$\pm$2.75 &
50.65 \scriptsize$\pm$2.83 &
22.35 \scriptsize$\pm$2.36 &
18.35 \scriptsize$\pm$2.18 &
40.65 \scriptsize$\pm$2.77 &
58.73 \scriptsize$\pm$2.79 &
19.81 \scriptsize$\pm$2.25 \\

\textbf{RAFT} &
18.35 \scriptsize$\pm$2.18 &
40.94 \scriptsize$\pm$2.78 &
51.93 \scriptsize$\pm$2.82 &
27.60 \scriptsize$\pm$2.52 &
16.47 \scriptsize$\pm$2.11 &
41.65 \scriptsize$\pm$2.79 &
53.22 \scriptsize$\pm$2.82 &
27.60 \scriptsize$\pm$2.52 \\

\textbf{PA-RAG} &
18.47 \scriptsize$\pm$2.18 &
38.59 \scriptsize$\pm$2.74 &
43.78 \scriptsize$\pm$2.81 &
32.29 \scriptsize$\pm$2.64 &
20.00 \scriptsize$\pm$2.26 &
42.59 \scriptsize$\pm$2.79 &
53.65 \scriptsize$\pm$2.82 &
29.17 \scriptsize$\pm$2.57 \\

\textbf{Chat-Vector} &
23.52 \scriptsize$\pm$2.39 &
40.47 \scriptsize$\pm$2.77 &
51.07 \scriptsize$\pm$2.83 &
27.60 \scriptsize$\pm$2.52 &
23.29 \scriptsize$\pm$2.38 &
44.94 \scriptsize$\pm$2.81 &
61.80 \scriptsize$\pm$2.75 &
24.47 \scriptsize$\pm$2.43 \\

\textbf{DKL} &
\textbf{24.47 \scriptsize$\pm$2.43} &
\textbf{46.12 \scriptsize$\pm$2.82} &
\textbf{57.94 \scriptsize$\pm$2.79} &
\textbf{31.77 \scriptsize$\pm$2.63} &
\textbf{25.41 \scriptsize$\pm$2.45} &
\textbf{48.94 \scriptsize$\pm$2.83} &
\textbf{65.24 \scriptsize$\pm$2.69} &
\textbf{29.17 \scriptsize$\pm$2.57} \\

\bottomrule
\end{tabular}
}
\caption{Comparing \ourmethodshort\ with baselines using different model architectures and sizes on Redbook1.}
\label{tab:arch:abl:appendix}
\end{table*}
The goal of this experiment is to establish that our proposed method \ourmethodshort\ works across model architectures and sizes. 
Consequently, we experiment using two additional models, varying both the size and model family. Specifically, we train {\href{https://huggingface.co/HuggingFaceTB/SmolLM2-1.7B-Instruct}{\textit{SmolLM2-1.7B-Instruct}}} and 
{\href{https://huggingface.co/Qwen/Qwen3-0.6B}{\textit{Qwen3-0.6B}}}
on Redbook1. 
\Cref{tab:arch:abl:appendix} presents the results. 
We observe that \ourmethodshort\ outperforms all baselines across model sizes and architectures, demonstrating its robustness.

\section{Token Swapping Ablation}
\label{sec:token-swapping-ablation}

\begin{table}[]
\resizebox{\columnwidth}{!}{%
\begin{tabular}{@{}l|l|lll@{}}
\toprule
 & \textbf{QA} & \multicolumn{3}{c}{\textbf{RAG}} \\ \midrule
 & \textbf{} & \textbf{All} & \textbf{\begin{tabular}[c]{@{}l@{}}Ret. \\ Success\end{tabular}} & \textbf{\begin{tabular}[c]{@{}l@{}}Ret. \\ Failure\end{tabular}} \\ \midrule
Instruct & 18.35 & 40.65 & 58.73 & 19.81 \\
DKL-e & 11.76 & 46.35 & 60.94 & 28.65 \\
\textbf{DKL} & \textbf{25.41} & \textbf{48.94} & \textbf{65.24} & \textbf{29.17} \\
\bottomrule
\end{tabular}
}
\caption{Effectiveness of using instruct LLM's embeddings during training. Comparing \ourmethodshort\ with a version trained directly on top of base LLM (e-\ourmethodshort) for RedBook1 using Qwen3-0.6B}
\label{tab:ablation-embed-swap-qwen}
\end{table}

First, we present the ablation results with Qwen3-0.6B in \cref{tab:ablation-embed-swap-qwen} and observe similar trends as observed with Mistral.

Next, we conducted experiments to determine where the major performance boost originates during embedding swap: from the OOD tokens in the base model, such as those introduced in the chat template, or from the tokens already trained in the base model. We use two strategies to automatically select the probable OOD tokens:
\begin{enumerate}
    \item  top-k: Select top-k tokens w.r.t. the L-2 norm of the difference between instruct and base models' token embeddings
    \item top-p: A nucleus sampling-like approach where we select the top-p tokens upto a threshold of 0.9.
\end{enumerate}

\begin{table}[ht!]
\resizebox{\columnwidth}{!}{%
\begin{tabular}{@{}l|r|rrr@{}}
\toprule
\multicolumn{1}{c|}{\textbf{}} & \multicolumn{1}{c|}{\textbf{QA}} & \multicolumn{3}{c}{\textbf{RAG}} \\ \midrule
 & \multicolumn{1}{l|}{} & \multicolumn{1}{c}{\textbf{All}} & \multicolumn{1}{c}{\textbf{\begin{tabular}[c]{@{}c@{}}Ret.\\ Success\end{tabular}}} & \multicolumn{1}{c}{\textbf{\begin{tabular}[c]{@{}c@{}}Ret.\\ Failure\end{tabular}}} \\ \midrule
\textbf{\ourmethodshort} & 73.80 & 86.58 & 92.13 & 79.26 \\
\textbf{top\_k-50-\ourmethodshort} & 74.12 & 87.22 & 94.38 & 77.78 \\
\textbf{top\_p-0.9-\ourmethodshort} & 74.12 & 84.66 & 91.01 & 76.30 \\
\textbf{bottom\_k-50-\ourmethodshort} & 72.52 & 83.39 & 92.13 & 71.85 \\
\textbf{bottom\_p-0.9 \ourmethodshort} & 73.48 & 81.79 & 88.20 & 73.33 \\
\textbf{e-\ourmethodshort} & 72.76 & 84.57 & 92.13 & 73.13 \\ \bottomrule
\end{tabular}
}
\caption{Ablation to study the effect of embeddings. top\_*-50/bottom\_*-50 refer to the embeddings with the most/least distance between the instruct and base models as chosen by the sampling method mentioned above. e-KnitLM refers to the run without doing any embedding swaps. }
\label{tab:ablation-emb-importance}
\end{table}

Row 2 and 3 (top\_k-50 and top\_p-0.9) represents the version where only the most different embeddings were swapped, and Row 4 and 5 (bottom\_k-50 and bottom\_p-0.1) represents the version where the most different embeddings were excluded, and only the rest of the embeddings were swapped.
Recall that e-\ourmethodshort (Row 6) represents the version where none of the embeddings were swapped.

We see that clearly the top-50 embeddings influence the final trained model much more than the rest, achieving performance equivalent to the full \ourmethodshort\ regime with just the top 50 embeddings swapped.
On the other hand, we can see Row 4 (bottom\_k-50-\ourmethodshort ) and Row 6 (e-\ourmethodshort) having comparable performance indicating that the embeddings which were the same in both the instruct and base models have little impact on the final performance.
Another interesting observation is the drop in performance between rows 2(top\_k-50-\ourmethodshort) and 3(top\_p-0.9).
One would expect that adding more meaningful tokens from the instruct model would improve performance, this is not what we see.
Our hypothesis as to why this happens can be explained due to 2 major phenomena:

\begin{itemize}
    \item the inherent value of each token from the instruct model i.e. how useful a particular token is to the training.
    \item the consistency of the entire embedding layer i.e., how consistent the embedding layer is with respect to its tokens (an embedding layer with only tokens from one model is said to be highly consistent where as a layer with 50\% tokens from one model and the rest from another is said to be highly inconsistent).
\end{itemize}

We posit that the OOD tokens help to a certain extent but soon start interfering with the other tokens as inconsistency within the layer increases. However, swapping the entire layer preserves consistency(as all tokens in the base model are swapped with instruct model’s) as well as utilising the more useful tokens during training. As swapping does not introduce any computational bottleneck we advise to always switch the entire layer as opposed to targeted tokens.



\section{Performance on general tasks}
\label{sec:general-task-performance}
We compare \ourmethodshort\ with Llama 3.1 8B Instruct (Instruct), RAFT, and PARAG on several benchmarks:

\begin{itemize}
    \item \textbf{Big Bench Hard} \citep{DBLP:journals/corr/abs-2210-09261}: 23 challenging tasks spanning language understanding and reasoning.
    \item \textbf{GPQA} \citep{rein2024gpqa}: Google-Proof Graduate-level STEM questions.
    \item \textbf{MATH-Hard} \citep{hendrycks2021measuring}: Difficult math competition questions.
    \item \textbf{MMLU-Pro} \citep{wang2024mmlupro}: 12k questions across diverse fields, measuring general knowledge.
    \item \textbf{MUSR (Multistep Soft Reasoning)} \citep{sprague2024musr}: Evaluates reasoning capabilities of LLMs.
\end{itemize}

\cref{tab:regression_table} reports the performance for RedBook 1.

\begin{table*}[!h]
\resizebox{\textwidth}{!}{%
\begin{tabular}{@{}lcccccc@{}}
\toprule
 & \textbf{Big Bench Hard} & \textbf{GPQA} & \textbf{MATH-Hard} & \textbf{MMLU Pro} & \textbf{MUSR} & \textbf{Aggregate} \\ \midrule
Instruct & 29.88 & 5.36 & 17.47 & 37.83 & 8.73 & 19.85 \\
\raft & 29.75 & 6.22 & 14.87 & 37.67 & 6.01 & 18.90 \\
\parag\ & 30.28 & 4.85 & 14.63 & 36.95 & 6.73 & 18.69 \\
\ourmethodshort\ & 29.69 & 7.59 & 17.11 & 38.08 & 6.75 & 19.84 \\ \bottomrule
\end{tabular}
}
\caption{General Task Performance}
\label{tab:regression_table}
\end{table*}
\ourmethodshort\ maintains competitive performance across all general benchmarks, while RAFT and PARAG show regression on general tasks relative to Instruct.

\section{Prompts and Examples}
\label{appendix:prompts}

This appendix presents the prompts used for three purposes: (i) filtering low-quality QA pairs from the dataset, (ii) evaluating responses generated by LLMs, and (iii) generating synthetic QA pairs. We also provide examples of QA pairs that were removed during the filtering process, along with sample responses from our method and the baseline models.

\subsection{Filtering Prompt}
The following prompt was used to identify Ill-formed QA pairs during dataset filtration. The filtering judge receives a question and its gold answer as input. It considers multiple criteria such as accuracy, relevance, clarity and usefulness and outputs a score from 1--10.

\begin{tcblisting}{
    colback=gray!5,
    colframe=gray!80,
    title=Filtering Prompt,
    listing only,
    breakable,
    listing options={
        basicstyle=\ttfamily\small,
        breaklines=true
    }
}
Rate each question-answer pair on a scale from 1-10, based on:
- Accuracy (0-3): factual correctness
- Relevance (0-2): relevance to content
- Clarity (0-2): clear language
- Usefulness (0-3): value for model learning

YOU MUST RETURN A VALID JSON OBJECT OR ARRAY WITH THIS EXACT SCHEMA:
{{
  "question": "Exact question text",
  "answer": "Exact answer text",
  "explanation": {{
    "Accuracy": "Short explanation of factual correctness",
    "Relevance": "Short explanation of relevance",
    "Clarity": "Short explanation of clarity",
    "Usefulness": "Short explanation of usefulness"
  }},
  "Accuracy": 2,
  "Relevance": 2,
  "Clarity": 2,
  "Usefulness": 2,
  "rating": 8
}}

OR FOR MULTIPLE PAIRS:
[
  {{
    "question": "Q1",
    "answer": "A1",
    "explanation": {{
      "Accuracy": "Explanation for Accuracy",
      "Relevance": "Explanation for Relevance",
      "Clarity": "Explanation for Clarity",
      "Usefulness": "Explanation for Usefulness"
    }},
    "Accuracy": 2,
    "Relevance": 2,
    "Clarity": 2,
    "Usefulness": 2,
    "rating": 8
  }},
  {{
    "question": "Q2",
    "answer": "A2",
    "explanation": {{
      "Accuracy": "Explanation for Accuracy",
      "Relevance": "Explanation for Relevance",
      "Clarity": "Explanation for Clarity",
      "Usefulness": "Explanation for Usefulness"
    }},
    "Accuracy": 3,
    "Relevance": 2,
    "Clarity": 2,
    "Usefulness": 2,
    "rating": 9
  }}
]

*** YOUR RESPONSE MUST BE VALID JSON AND NOTHING ELSE - NO EXPLANATION, NO MARKDOWN ***

QA pairs to rate:
{pairs}
\end{tcblisting}

\subsection{LLM-as-a-Judge Prompt}
The following prompt was used to evaluate model-generated responses. The model is provided with the question, gold answer and model generated answer, and it outputs a binary rating (0/1) according to the specified evaluation rules.

\begin{tcblisting}{
    colback=gray!5,
    colframe=gray!80,
    title=LLM Evaluation Prompt,
    listing only,
    breakable,
    listing options={
        basicstyle=\ttfamily\small,
        breaklines=true
    }
}
You are an evaluator. Your task is to compare a Ground-truth Answer and a Prediction to decide if the Prediction correctly answers the given Question.

Evaluation Rules:
(1) Correctness: A correct prediction must include all essential information from the Ground-truth Answer. Extra information is allowed if it does not contradict the Ground-truth. If the Prediction states something as a possibility, treat it as a definitive statement.

(2) Function, Tool Names, and API Calls: If the Ground-truth Answer contains specific function names, tool names, API calls, or exact command identifiers, the Prediction must contain the same identifier(s) or clearly equivalent forms. Minor syntactic or formatting variations that do not change meaning should be treated as equivalent. For example, leading flag prefixes such as -, --, or no prefix at all when they clearly refer to the same option;  underscore vs hyphen differences in identifiers when the intent is identical;  surrounding punctuation or formatting differences such as backticks, quotes, parentheses, or code block notation; small whitespace differences or capitalization differences that do not change the identifier's meaning etc. However, replacements that change the actual function/tool/API name, or substitute a different command that would change the behavior are considered incorrect. Do not penalize a prediction if it contains additional function / tool / API names as long as the ones present in the Ground-Truth are covered.
	
(3) URLs: If the Ground-truth Answer contains specific URLs, the Prediction should reference the same URL or an equivalent canonical form. Minor differences that do not change the target resource (for example, presence or absence of a trailing slash, or http vs https when both resolve to the same canonical resource) should be treated as equivalent. Altering the domain, path, or query such that the resource is different is incorrect.

Scoring Rules:
If the Prediction is correct according to the above rules, output <score>1</score>. If the Prediction is incomplete or incorrect, output <score>0</score>.

Output Format:
<explanation>
...
</explanation>
<score>
...
</score>

First provide reasoning inside <explanation> and </explanation> tags. Then output the score as specified above within <score> and </score> tags. Do not include any extra text outside these tags.

\end{tcblisting}

\subsection{Prompt for generating synthetic QA}
The prompt generates fully contextualized question–answer pairs from a document, covering the entire content and formatted with specific tags.

\begin{tcblisting}{
    colback=gray!5,
    colframe=gray!80,
    title=QA Generation Prompt,
    listing only,
    breakable,
    listing options={
        basicstyle=\ttfamily\small,
        breaklines=true
    }
}
Create question answer pairs from the document given below within <document> tags. Title of the document is given in the first line of the document. Do not use co-referencing and pronouns at all in the questions. Do not refer to the document in the question like "according to the document ..." or any similar paraphrasing. When needed, contextualize the question by using the topic that the question is about. You can use the title of the document as well for contextualizing. There are several figures in the document, while referring to the figure in any question, contextualize it by mentioning the title of the passage it was present in. Put questions within <question> and </question> tags and answers within <answer> and </answer> tags. Ensure that the question and answers cover the entire document. When you are done generating QA pairs, generate </done> token.
\end{tcblisting}

\subsection{Examples of QA Pairs Removed During Filtering}
Below are three representative examples of QA pairs that were filtered out by the LLM-as-a-Judge. Each example shows the question and gold answer.

\definecolor{lightred}{rgb}{0.99,0.95,0.95}

\newtcolorbox{qabox}[1]{
    colback=lightred,
    colframe=red!30,
    boxrule=0.5pt,
    arc=2pt,
    left=6pt,
    right=6pt,
    top=6pt,
    bottom=6pt,
    enhanced,
    overlay={
        \node[draw=red!30, fill=red!30, rounded corners=2pt, inner sep=2pt, 
              font=\bfseries, anchor=north east] 
        at (frame.south east) 
        {\textcolor{red}{\Large$\times$} #1};
    }
}

\begin{qabox}{Vague}
\textbf{Question:} What is the state of the second volume to be mapped to the hostcluster?\\
\textbf{Gold Answer:} The state of the second volume to be mapped to the hostcluster is present.
\end{qabox}

\vspace{12mm}

\begin{qabox}{Incomplete}
\textbf{Question:} What was the status, memory, and CPU usage of the voting-app-worker-py-1 pod in the Red Hat OpenShift Cluster on IBM LinuxONE?\\
\textbf{Gold Answer:} The status, memory, and CPU usage of the voting-app-worker-py-1 pod in the Red Hat OpenShift Cluster on IBM LinuxONE were as follows: \\
- Status: Completed \\
- Memory: Not specified \\
- CPU: Not specified
\end{qabox}

\vspace{12mm}

\begin{qabox}{Contextual}
\textbf{Question:} What are the settings for the infra nodes in the deployment of Red Hat OpenShift on IBM Z?\\
\textbf{Gold Answer:} The infra nodes have settings such as disk size, RAM, vCPU, VM name, IP address, and hostname. These settings are defined in Section 10 of the all.yaml file.
\end{qabox}

\definecolor{lightgreen}{rgb}{0.95,0.99,0.95} 
\definecolor{lightred}{rgb}{0.99,0.97,0.97}   

\newtcolorbox{redbox}[1]{
    colback=lightred,
    colframe=red!50,
    boxrule=0.5pt,
    arc=2pt,
    left=6pt,
    right=6pt,
    top=6pt,
    bottom=6pt,
    enhanced,
    overlay={
        \node[draw=red!50, fill=red!50, rounded corners=2pt, inner sep=2pt, 
              font=\bfseries, anchor=north east] 
        at (frame.south east) 
        {\textcolor{red}{\Large$\times$} #1};
    }
}

\newtcolorbox{greenbox}[1]{
    colback=lightgreen,
    colframe=green!50,
    boxrule=0.5pt,
    arc=2pt,
    left=6pt,
    right=6pt,
    top=6pt,
    bottom=6pt,
    enhanced,
    overlay={
        \node[draw=green!50, fill=green!50, rounded corners=2pt, inner sep=2pt, 
              font=\bfseries, anchor=north east] 
        at (frame.south east) 
        {\textcolor{green!80!black}{\Large$\checkmark$} #1};
    }
}

\definecolor{lightgold}{rgb}{1.0, 0.98, 0.88} 

\newtcolorbox{goldbox}{
    colback=lightgold,
    colframe=orange!60!black,
    boxrule=0.5pt,
    arc=2pt,
    left=6pt,
    right=6pt,
    top=6pt,
    bottom=6pt,
    enhanced
}

\subsection{Example Model Responses}

\begin{goldbox}
\textbf{Question:} What command is used to upgrade the IBM Storage Virtualize collection to the latest version from Ansible Galaxy?

\textbf{Gold Answer:} The command used to upgrade the IBM Storage Virtualize collection to the latest version from Ansible Galaxy is: \\
\textit{ansible-galaxy collection install ibm.storage\_virtualize --force}
\end{goldbox}

\vspace{4mm}

\begin{redbox}{Instruct}
To upgrade the IBM Storage Virtualize collection to the latest version from Ansible Galaxy, you can use the following command:

```
\textit{ansible-galaxy collection install ibm.storwize:latest}
```
\end{redbox}

\vspace{4mm}

\begin{redbox}{RAFT}
The command used to upgrade the IBM Storage Virtualize collection to the latest version from Ansible Galaxy is:

```\textit{bash
ansible-galaxy collection install -r requirements.yml --force}
```

\end{redbox}

\vspace{4mm}

\begin{redbox}{PA-RAG}
The command to upgrade the IBM Storage Virtualize collection to the latest version from Ansible Galaxy is:

```\textit{bash
ansible-galaxy collection install -r requirements.yml}
```

\end{redbox}

\vspace{4mm}

\begin{greenbox}{\ourmethodshort}
"The command to upgrade the IBM Storage Virtualize collection to the latest version from Ansible Galaxy is:

\textit{ansible-galaxy collection install ibm.storage\_virtualize --force}"
\end{greenbox}

\vspace{4mm}

Only \ourmethodshort\ provides the exact command that correctly upgrades the IBM Storage Virtualize collection. The instruct's response contains a typo in the collection name (`storwize` instead of `storage\_virtualize`), while RAFT and PA-RAG incorrectly rely on a requirements file, which is not specified in the ground truth.

\section{LLM Usage}
During the preparation of this manuscript, we employed a Large Language Model (LLM) as a writing support tool. Specifically, LLM was used to polish the phrasing, improve grammatical accuracy, and provide paraphrased alternatives to enhance clarity and readability. The LLM’s role was limited to language refinement, and all suggested edits were reviewed and verified by the authors before inclusion.

\end{document}